\documentclass[sigconf, nonacm]{acmart}

\newcommand\vldbdoi{XX.XX/XXX.XX}

\newcommand\vldbavailabilityurl{xxx}
 
\usepackage{threeparttable}
\usepackage{booktabs}
\usepackage{bm}
\usepackage{amsmath}

\usepackage{amssymb,amsfonts}
\usepackage[ruled,linesnumbered, vlined]{algorithm2e}
\usepackage{algorithmic}
\usepackage{caption}
\usepackage{bigstrut}
\usepackage{subfig}
\usepackage{enumitem}
\usepackage{float}
\usepackage{balance}
\usepackage{multirow}
\usepackage{ulem}
\usepackage{tabularx}
\usepackage{bbm}
\usepackage{xcolor}
\usepackage{makecell}
\usepackage{array}
\usepackage{booktabs}

\theoremstyle{definition}

\newtheorem{prop}{Proposition}

\newcommand{\M}[1]{\mathcal{M}}
\AtBeginDocument{%
  \providecommand\BibTeX{{%
    \normalfont B\kern-0.5em{\scshape i\kern-0.25em b}\kern-0.8em\TeX}}}

\begin{document}

\title{UniCL: A Universal Contrastive Learning Framework for Large Time Series Models}

%


\author{Jiawei Li}
\affiliation{%
	\institution{HKUST (GZ)}
	\city{Guangzhou, China}
	\country{}
}
\email{jli226@connect.hkust-gz.edu.cn}

\author{Jingshu Peng}
\affiliation{%
	\institution{HKUST}
	\city{Hong Kong SAR, China}
	\country{}
}
\email{jpengab@cse.ust.hk}

\author{Haoyang Li}
\affiliation{%
	\institution{HKUST}
	\city{Hong Kong SAR, China}
	\country{}
}
\email{hlicg@connect.ust.hk}

\author{Lei Chen}
\affiliation{%
	\institution{HKUST (GZ) \& HKUST}
	\city{Hong Kong SAR, China}
	\country{}
}
\email{leichen@cse.ust.hk}

\begin{abstract}

Time-series analysis plays a pivotal role across a range of critical applications, from finance to healthcare, which involves various tasks, such as forecasting and classification. 
To handle the inherent complexities of time-series data, such as high dimensionality and noise,
traditional supervised learning methods first annotate extensive labels for time-series data in each task, which is very costly and impractical in real-world applications.
In contrast, pre-trained foundation models offer a promising alternative by leveraging unlabeled data to capture general time series patterns, which can then be fine-tuned for specific tasks. 
However, existing approaches to pre-training such models typically suffer from  \textit{high-bias} and \textit{low-generality} issues due to the use of predefined and rigid augmentation operations and domain-specific data training. 
To overcome these limitations, this paper introduces {UniCL}, a universal and scalable contrastive learning framework designed for pretraining time-series foundation models across cross-domain datasets.
Specifically, we propose a unified and trainable time-series augmentation operation to generate pattern-preserved, diverse, and low-bias time-series data by leveraging spectral information. 
Besides, we introduce a scalable augmentation algorithm capable of handling datasets with varying lengths, facilitating cross-domain pretraining.
Extensive experiments on two benchmark datasets across eleven domains validate the effectiveness of UniCL, demonstrating its high generalization on time-series analysis across various fields.

\end{abstract}

\keywords{Foundation Model, Time Series, Data Management}
\twocolumn
\maketitle

\section{introduction}\label{sec:intro}

Time series data, which consists of sequences of real values recorded over time intervals, are prevalent in numerous real-world applications~\cite{wen2022robust}, including finance, healthcare, environmental monitoring, and manufacturing. 
Time series analysis involves various tasks such as forecasting~\cite{sezer2020financial} and  classification~\cite{ismail2019deep},
	which are crucial for decision-making processes.
However, analyzing time series data remains challenging due to their inherent properties, such as high dimensionality, noise, non-stationarity, periodicity, etc.~\cite{trirat2024universal}.
 These features pose difficulties in effectively capturing and leveraging the underlying patterns in time series data, thus impacting its practical applications~\cite{ma2023survey}, highlighting the ongoing challenge of analyzing time series data across various tasks.

In general, existing time-series analysis in deep learning can be categorized into two types, supervised learning (SL)-based approaches~\cite{salehinejad2017recent,wen2022transformers} and pretrained foundation model (PFM)-based approaches~\cite{jin2023large, zhang2024self}. 
First, SL-based approaches~\cite{wu2021autoformer, liu2021pyraformer,zhou2021informer,koh2021deep} propose to train a deep learning model, such as transformers~\cite{zhou2022fedformer,wu2021autoformer,Yuqietal-2023-PatchTST} and CNNs~\cite{zhao2017convolutional}, on labeled time series data. However, these SL-based approaches rely on a large number of annotated time series data~\cite{khaertdinov2021contrastive}, which is often impractical in many real-world scenarios where obtaining labeled time-series data is costly or infeasible. 
Second, current researchers ~\cite{dong2024simmtm,yue2022ts2vec,liu2023unitime,bian2024multi} propose to first pre-train a foundation model (e.g., large language models) on time-series data, such as those initially large language models (e.g., ChatGPT~\cite{Thornburg01}) developed for processing natural language. 
The basic idea is to first pre-train the foundation model to capture the general intrinsic patterns of time-series data.
Subsequently, the foundation model can be fine-tuned on specific time-series tasks using a smaller set of labeled data. 
These PFM-based approaches can  leverage the learned general patterns to enhance performance across a variety of tasks.

Depending on the technique employed to pre-train the foundation model, existing research~\cite{li2023ti, zerveas2021transformer,dong2024simmtm} first propose to use mask-based approaches to pre-train the foundation model in time series data. 
They directly use the foundation model to predict the next values in a time series~\cite{bian2024multi,liu2023unitime,rasul2023lag} or reconstruct randomly masked values of the data~\cite{dong2024simmtm}, \textit{i.e.}, predictive objectives. 
However, these mask-based pre-training approaches require numerous unlabeled time-series data to enable the foundation model to capture general time-series patterns~\cite{wen2022transformers}. 
Unlike corpus data, the time-series are scarce in the open-world website~\cite{ansari2024chronos,liu2024auto}. Consequently, these mask-based approaches achieve suboptimal performance on the downstream tasks.
Recently, to alleviate the heavy reliance on numerous time series data, contrastive learning approaches~\cite{yeh2023toward,ijcai2021-324,meng2023mhccl,woo2022cost,luo2023time} are proposed to augment time-series data and pre-train the foundation model based on augmented data. 
The basic idea is to first generate positive views for each time series data  to preserve its key patterns. 
Then, the foundation model is optimized by maximizing  the
representation similarity between positive view pairs and  minimizing the representation similarity between
positive views and other randomly selected time series data (i.e., negative views), known as contrastive objectives.
This enables the pre-trained model to distinguish patterns across different time-series data, thereby facilitating downstream tasks~\cite{ma2023survey}.

Nevertheless, existing CL-based pretrained foundation models suffer from two issues, i.e., \textit{high-bias} and \textit{low-generality} issues. 
First, these models typically predefine and apply a set of time series augmentation operations, such as permutation~\cite{meng2023mhccl,poppelbaum2022contrastive}, random masking~\cite{yue2022ts2vec,wang2024contrast},
and warping~\cite{fan2020self} to generate positive views. 
However, existing predefined operations ignore the time-series intrinsic properties,
such as periodicity~\cite{qiu2024tfb},
thereby introducing  high bias and noise. For instance, permutation shuffles the sequential information in time series data, potentially resulting in the loss of its inherent patterns.
Consequently, the foundation models, optimized by maximizing the similarity between the generated positive views, may not capture the patterns of real time-series data, thus degrading the performance of downstream tasks~\cite{Zheng2023SimTSRC, yue2022ts2vec, ijcai2021-324}. 
Secondly, these models are generally pretrained within a single specific domain~\cite{yue2022ts2vec,woo2022cost,luo2023time}, such as medicine~\cite{wang2024contrast}. However, time series data can vary significantly across different domains (e.g.,  medical, traffic, and weather) in aspects of the number of variables, sequence length, and frequency (e.g., daily, weekly, monthly).
Due to these variations, foundation models trained on one domain often fail to perform effectively across other domains~\cite{ma2023survey}. 
In other words, existing approaches~\cite{wang2024contrast,yue2022ts2vec,woo2022cost} need to pre-train a single foundation model for each specific domain, which is time-consuming and very costly.

To address the \textit{high-bias} and \textit{low-generality} issues, we propose a universal pre-trained foundation model for time series analysis. 
Firstly, to address the \textit{high-bias} issue, we propose to use effective pattern-preserved augmentation operations to generate positive views for the time series data.
By optimizing the foundation model on these augmented views, it becomes capable of learning and retaining the intrinsic patterns present in real-world time series data, thereby reducing bias.
Secondly, to overcome \textit{low-generality} issue, we propose to pre-train the foundation model across a diverse array of time series data from various domains.
In such a way, this foundation model can learn a wide range of time series patterns specific to different domains. 
As a result, the foundation model can generalize on various domains and thereby facilitate their downstream tasks.
However, there are three technique challenges to achieving this unified foundation model.

\begin{itemize}[leftmargin=*]
	\item 
	There is no theoretical analysis and established metrics for preserving the patterns of time-series data with deep learning. Without these metrics, we cannot design effective augmentation operations to generate positive views that keep the intrinsic patterns.
	
	\item
	Due to the high variations (e.g., variable number and sequence length) in time-series data from different domains, it is challenging to design a scalable and unified augmentation algorithm that is applicable across these diverse settings.

	\item
	Last but not least, existing studies train the LLM-based  encoder for time-series analysis by optimizing the predictive objective, yet, the exploration of the contrastive objective has received comparatively less attention.
		
\end{itemize}

To address the above technique challenges, we propose {UniCL}, a universal contrastive learning framework designed for pre-training time-series foundation models across diverse domains.
{Firstly}, we empirically reveal a positive correlation between the bias of time series embeddings and the spectral distance between augmented and raw series. Building on this insight, we propose a unified and trainable time series augmentation operation with theoretical guarantees. Through optimization of two novel losses, our proposed operation generates spectrum-preserved, diverse, and low bias augmented series. 
{Secondly}, to tackle high variations in datasets and enable large-scale automatic pre-training, we propose a scalable and unified augmentation algorithm. This algorithm utilizes spectrum-preserved time-series segmentation, augmenting each subseries individually. We demonstrate that the difference in convergence loss between the scalable and non-scalable algorithms can be bounded. 
 {Thirdly}, to fully leverage the potential of the LLM backbone, we train a transformer-based encoder on 40 cross-domain datasets, initialized with pre-trained weights from the text encoder of CLIP~\cite{radford2021learning}, owing to our shared contrastive objectives.
We summarize the novel contributions of this paper as follows.


\begin{itemize}[leftmargin=*]
	\item 
	We present UniCL, an end-to-end general framework for pre-training large foundation time-series models based on contrastive learning, capable of handling high variation time-series datasets.
	
	\item
	We reveal the factor of representation bias based on a novel metric, and propose a unified and trainable augmentation operation with theoretical guarantees. We then propose two novel losses to facilitate the optimization of this operation.
	
	\item
	We propose a scalable and unified algorithm to handle data with varied length by pattern-preserved segmentation and concatenation, and demonstrate bounded convergence loss differences between scalable and non-scalable algorithms.

	\item
	To the best of our knowledge, we are the first to train the LLM backbone with contrastive objectives for general time-series analysis. We train the UniCL on 40 cross-domain datasets, and provide a comprehensive evaluation of its performance across two downstream tasks.

	
\end{itemize}



\section{Preliminaries and related works}\label{sec:preliminary}

\begin{table}[t]
	\setlength{\abovecaptionskip}{0.2cm} 
	\setlength{\belowcaptionskip}{-0.2cm}
	\caption{Summary of important notations}
	\label{tab:notations}
	\begin{tabular}{c|l}
		\toprule 
		Notation & Description \\ 
		\midrule
		$t, T$ & General time index \\ \hline

		
		$n$ & The number of variables \\\hline
		$\mathbf{x} \in \mathbb{R}^{T}$ & The observed sequences of variable \\ \hline
		$x_i \in \mathbb{R}$ & The $i$-th value of $\mathbf{x}$ \\ \hline
		$\mathbf{X}^{(j)} \in \mathbb{R}^{n\times T}$ & The $j$-th observed time series data \\ \hline
		
		$\mathbf{x}^{(j,i)}$ & The $i$-th variable in $\mathbf{X}^{(j)}$ \\ \hline

		$\mathbf{X}_{t:t+H}$ & Values of $\mathbf{X}$ between time $t$ and $t+H$ \\ \hline
		
		${f}_\theta$ & The encoder model for time series data \\ \hline
		
				${f}_{\theta^\dagger}$ &  Optimized foundation  model on time-series data \\ \hline

			$q_\varphi$ & The decoder model for prediction \\ 
		
		\bottomrule 
	\end{tabular}
\end{table}

In this section, we first introduce the basic concepts in time-series data analysis and then 
introduce the foundation models for time-series data analysis. 
The important notations are listed in Tab.~\ref{tab:notations}.

\subsection{Time-series Data Analysis}
\label{ssec:tasks}
Time series data are a set of sequences of observations where each sequence corresponds to a different variable, and all sequences are recorded over the same time periods.
Time series data plays a crucial role in various fields~\cite{wen2022robust}, such as economics, finance, environmental science, and engineering.
Formally, a time series data instance with $n$ variables over time $T$ can be denoted as $\mathbf{X}=[\mathbf{x}^{(i)}]_{i=1}^{n} \in \mathbb{R}^{n \times T}$, where $\mathbf{x}^{(i)}= [x^{(i)}_1, x^{(i)}_2, \cdots, x^{(i)}_T] \in \mathbb{R}^{T}$ in the observed real value of variable $x^{(i)}$ from time $1$ to $T$. 
In general, if the number of variable $n$ is 1 (resp, $n\ge2$), this time-series data is called univariate (resp. multivariate) time-series data. 
Time-series analysis involves various tasks such as forecasting~\cite{lim2021time} and classification~\cite{ismail2019deep}.
We first give a general time-series data analysis problem definition for various tasks.
Formally, given a training time-series data $D_{task}=\{\mathbf{X}^{(j)}, \mathbf{Y}^{(j)}\}_{j=1}^m$ with $m$ instances for a task, where $\mathbf{X}^{(j)}=[\mathbf{x}^{(j,i)}]_{i=1}^{n}$ denotes each time series instance and $\mathbf{Y}^{(j)}$ is the label of  $\mathbf{X}^{(j)}$, 
the target is to supervised train an encoder model $f_\theta$ and a simple decoder model $q_{\varphi}$ {\color{black}(e.g., simple linear regression model~\cite{yue2022ts2vec})} on data $D_{task}$. 
The encoder model $f_\theta$  is used to encode each time-series data $\mathbf{X}^{(j)}$ to
an embedding matrix with $D$ dimensions $\mathbf{Z}^{(j)} = f_\theta(\mathbf{X}^{(j)}) \in \mathbb{R}^{n \times D}$. 
Then the decoder $q_\varphi$ will use the $\mathbf{Z}^{(j)}$ to predict the task labels $\mathbf{\hat{Y}}^{(j)}=q_{\varphi}(\mathbf{Z}^{(j)})$. 
In general, the parameters of the encoder model $f_\theta$ and the decoder model $q_{\varphi}$  can be optimized by
 minimizing the task loss $\mathcal{L}_{task}(\cdot)$ as follows:
 \begin{align}\label{eq:unfied_loss}
\theta^*,\varphi^* = \arg\min_{\theta,\varphi}\mathcal{L}_{task}(f_\theta, q_\varphi,
D_{task})=\sum_{j=1}^ml_{task}(\mathbf{\hat{Y}}^{(j)}, \mathbf{Y}^{(j)}).
 \end{align}
Then, we discuss how to adapt this definition for each specific task as follows.


\begin{itemize}[leftmargin=*]
	\item \textbf{Time-series Classification Task.}
	The target is to predict the label $y^{(j)} \in \mathcal{Y}$ of each time-series data instance $\mathbf{X}^{(j)}$. 
	In general, $D_{task}=\{\mathbf{X}^{(j)}, \mathbf{y}^{(j)}\}_{j=1}^{m}$, where $\mathbf{y}^{(j)} \in \{0,1\}^{\mathcal{Y}}$ is  multiple-label ground  truth of $\mathbf{X}^{(j)}$. 
	Based on cross-entropy, the classification loss $\mathcal{L}_{cl}(\cdot)$~\cite{zhou2023onefitsall,sun2023test} can be defined  as: $\mathcal{L}_{cl}(f_{\theta}, q_\varphi, D_{task})=\frac{1}{m}\sum_{j=1}^{m}\sum_{i=1}^{|\mathcal{Y}|} \mathbf{y}^{(j)}[i]\log p_{ \mathbf{y}^{(j)}[i]}$, where $p_{ \mathbf{y}^{(j)}[i]}$ is the predicted probability of each label $ \mathbf{y}^{(j)}[i]$ for data $\mathbf{X}^{(j)}$.
	
	\item  \textbf{Time-series Forecasting Task.}
		The target is to predict the future values $\mathbf{X}_{t:t+H_f} \in \mathbb{R}^{n \times H_f}$ of a time-series data based on the previous observation $\mathbf{X}_{t-H_p:t}\in \mathbb{R}^{n \times H_p}$, where $H_p$ is the look back window size, and $H_f$ denotes the number of future prediction values. 
		In general,
		$D_{task}=\{\mathbf{X}^{(j)}_{t-H_p:t},\mathbf{X}^{(j)}_{t:t+H_f}\}_{j=1}^m$, 	and forecasting loss $\mathcal{L}_{fl}(\cdot)$~\cite{jin2023time,liu2023unitime} can be defined as $\mathcal{L}_{fl}(f_{{\theta}}, q_\varphi, D_{task})=\frac{1}{m}\sum_{j=1}^{m}\lVert\mathbf{X}^{(j)}_{t:t+H_f} - \mathbf{\hat{X}}^{(j)}_{t:t+H_f} \rVert_2^2$, where $\mathbf{\hat{X}}^{(j)}_{t:t+H_f}=q_\varphi(f_\theta(\mathbf{X}^{(j)}_{t-H_p:t}))$ is the predicted values by the encoder and decoder models.

\end{itemize}

To handle each specific task in time-series analysis, existing research~\cite{wu2021autoformer,liu2021pyraformer,zhou2021informer} propose to prepare the training data set. Then, they train deep learning models, such as transformers and convolutional neural networks (CNNs), on labeled time series data. 
{\color{black}However, these approaches need substantial labeled data, which is time-consuming and very costly~\cite{liu2024large}.}

\subsection{Foundation Models for Time-series Data}
To alleviate the reliance on labeled data, current researchers propose to learn a foundation model that captures the general patterns of time-series data and then fine-tune this foundation model with limited labels for the downstream tasks.

 \subsubsection{Pretrained Foundation Models}\label{sssec:foundation_model} 
	In general, existing  pretrained foundation models for time-series data analysis can be categories into the following three types, i.e., \textit{pretrained language model-based}, \textit{mask-based}, and \textit{contrastive learning (CL)-based} foundation models.
	
	   {\textit{\textbf{1) Pretrained Language Models.}}}
	   {\color{black}Recently, the  language models (LMs) pretrained on numerous corpus data, such as  BERT~\cite{devlin2018bert}, T5~\cite{2020t5}, LLaMA~\cite{Touvron2023LLaMAOA}, and GPT~\cite{radford2019language}, have demonstrated their strong text understanding ability in natural language processing tasks, such as question answering. 

	   These LMs employ transformer models with self-attention mechanisms to capture long-range dependencies in data, which have the potential to capture complex temporal dependencies in time-series data.
	   Therefore,
	   several researchers~\cite{gruver2023llmtime,liu2024lstprompt} directly take these pretrained LMs as the foundation model $f_{\theta^*}$ and propose to apply these pretrained LMs in the time-series data analysis tasks. 
	   The basic idea is to propose time-series data encoding approaches to transform the time-series data into a text-like format that can be processed by language models.
	   For instance, 
	   LLMTIME~\cite{gruver2023llmtime} encodes time-series as a string of numerical digits, with each digit separated by spaces.
	   LSTPrompt~\cite{liu2024lstprompt} and PromptCast~\cite{xue2023promptcast} incorporate domain information (\textit{e.g.}, traffic, weather) and frequency (e.g., hour, day) into template-based descriptions, embedding them using a text tokenizer to guide the LLM.
	   
	   However, LMs are inherently designed for handling text data, which is discrete and categorical, whereas time-series data is usually continuous and numeric. 
	   Therefore, existing LMs cannot effectively capture the patterns and semantics of time-series data, leading to suboptimal performance on time-series tasks.
	   To overcome this problem, current researchers propose mask-based~\cite{zerveas2021transformer,dong2024simmtm} and CL-based ~\cite{yue2022ts2vec,zhang2022self} foundation models pretrained on time-series directly for time-series tasks.}

	   {\textit{\textbf{2) Mask-based  Foundation Models.}}}
	%
%
Mask-based foundational models are typically pre-trained using tasks where specific values in the input time-series are masked, and the model is trained to predict them based on contextual information. Based on the prediction targets, existing approaches can be classified into reconstruction-based~\cite{zerveas2021transformer,li2023ti,dong2024simmtm,gao2024units} and prediction-based~\cite{bian2024multi,liu2024auto,jin2023time}. 
In reconstruction-based methods, a binary random mask $\mathbf{M}\in\{0,1\}^{n\times T}$ is generated to mask the input series as $\mathbf{X}^{(j)}\odot\mathbf{M}^{(j)}$. Then, the pretext task is to optimize the reconstruction loss~\cite{zerveas2021transformer}, which is defined as follows:
\begin{align}
\mathcal{L}_{rl}=\frac{1}{m}\sum_{j=1}^{m}\lVert\mathbf{X}^{(j)}\odot(1-\mathbf{M}^{(j)}) - \mathbf{\hat{X}}^{(j)}\odot(1-\mathbf{M}^{(j)}) \rVert_2^2,
\end{align} 
where $\mathbf{\hat{X}}^{(j)}$ is the predicted value.
In prediction-based methods, models are trained to predict future values directly, with a typical prediction loss~\cite{bian2024multi} defined as follows:
\begin{align}
	\mathcal{L}_{pl}=\frac{1}{m}\sum_{j=1}^{m}\lVert\mathbf{X}^{(j)}_{t:t+H_f} - \mathbf{\hat{X}}^{(j)}_{t:t+H_f} \rVert_2^2.
\end{align} 

However, masked modeling necessitates abundant data for effective representation learning due to its reliance on self-encoding reconstruction and prediction objectives~\cite{wen2022transformers}, and masked values are often easily predicted from neighboring time points~\cite{cheng2023timemae}. The availability of large-scale datasets are required to ensure that the model is exposed to diverse patterns and variations in the data.

   {\textit{\textbf{3) Contrastive Learning-based  Foundation Models.}}}
   Contrastive learning aims to train the encoder $f_{\theta}$ by contrasting between positive and negative samples, and augmentation contrast is one of the most widely used contrastive frameworks~\cite{zhang2023ssl4ts}. Specifically, given a family of augmentations $\mathcal{T}$, data augmentation operations $\tilde{t}^{(i)}, \hat{t}^{(i)}\sim\mathcal{T}$ are applied to create positive views $\tilde{\mathbf{X}}^{(j)}=[\tilde{\mathbf{{x}}}^{(j,i)}]_{i=1}^{n}$ and $\hat{\mathbf{X}}^{(j)}=[\hat{\mathbf{{x}}}^{(j,i)}]_{i=1}^{n}$, where $\tilde{\mathbf{{x}}}^{(j,i)}=\tilde{t}^{(i)}(\mathbf{x}^{(j,i)})$ and $\hat{\mathbf{{x}}}^{(j,i)}=\hat{t}^{(i)}(\mathbf{x}^{(j,i)})$. The positive pairs $(\tilde{\mathbf{{x}}}^{(j,i)}, \hat{\mathbf{{x}}}^{(j,i)})_{i=1}^{n}$ are expected to retain crucial temporal information, and the corresponding representations, $\tilde{\mathbf{z}}^{(j,i)} = f_{\theta}(\tilde{\mathbf{x}}^{(j,i)})$ and  $\hat{\mathbf{z}}^{(j,i)} = f_{\theta}(\hat{\mathbf{x}}^{(j,i)})$, should exhibit proximity within the embedding space. The negative views of $\tilde{\mathbf{{x}}}^{(j,i)}$, denoted as $Neg(\tilde{\mathbf{{x}}}^{(j,i)})$, comprise a set of time-series that exhibit dissimilarity to $\tilde{\mathbf{{x}}}^{(j,i)}$. A typical choice~\cite{yue2022ts2vec} is $Neg(\tilde{\mathbf{{x}}}^{(j,i)})=\{\{\tilde{\mathbf{{x}}}^{(j,k)}\}_{k\neq i} \cup \{\hat{\mathbf{{x}}}^{(j,k)}\}_{k=1}^{n} \}$. Similarly, $Neg(\hat{\mathbf{{x}}}^{(j,i)})=\{\{\hat{\mathbf{{x}}}^{(j,k)}\}_{k\neq i} \cup \{\tilde{{\mathbf{{x}}}^{(j,k)}}\}_{k=1}^{n} \}$. To promote proximal representations of positive pairs while ensuring distant representations of negative views, the encoder $f_{\theta}$ is optimized by the contrastive loss $l_{cl}(\cdot)$:
\begin{equation}
	\theta^{\dagger} = \arg \min_{\theta} \frac{1}{2mn}\sum_{j=1}^{m}\sum_{i=1}^{n}l_{cl}(f_{\theta},\tilde{\mathbf{{x}}}^{(j,i)},\hat{\mathbf{{x}}}^{(j,i)},Neg(\cdot))
\end{equation}

\noindent One widely used $l_{cl}$~\cite{chen2020simple} is defined as:
\begin{equation}
\begin{aligned}
	&l_{cl}(f_{\theta},\tilde{\mathbf{{x}}}^{(j,i)},\hat{\mathbf{{x}}}^{(j,i)},Neg(\cdot))\\
	& = - \log \frac{\exp(sim(\tilde{\mathbf{z}}^{(j,i)}, \hat{\mathbf{z}}^{(j,i)})/\tau)}
	 {\sum_{\mathbf{x}\in Neg(\tilde{\mathbf{{x}}}^{(j,i)})} 
	 \exp(sim(\tilde{\mathbf{z}}^{(j,i)}, f_{\theta}(\mathbf{x}))/\tau)}\\
	&  - \log \frac{\exp(sim(\hat{\mathbf{z}}^{(j,i)}, \tilde{\mathbf{z}}^{(j,i)})/\tau)}
{\sum_{\mathbf{x}\in Neg(\hat{\mathbf{{x}}}^{(j,i)})} 
	\exp(sim(\hat{\mathbf{z}}^{(j,i)}, f_{\theta}(\mathbf{x}))/\tau)}\\
\end{aligned}
\end{equation}
\noindent Where $\tau$ represents the temperature parameter, and $sim(\cdot)$ denotes the similarity function.

	\begin{algorithm}[t]

		\KwIn{$D_{train}=\{\{\mathbf{X}^{(j)}\}_{j=1}^m\}$, $D_{task}=\{\{\mathbf{X}^{(j)}\}_{j=1}^{m'},\{ \mathbf{Y}^{(j)}\}_{j=1}^{m'}\}$, the augmentation family $\mathcal{T}$, the encoder $f_\theta$, the decoder $q_{\varphi}$, epoch number $T$}
		\KwOut{ The encoder $f_{\theta^*}$ and the decoder $q_{\varphi^*}$}
		\tcp{Contrastive learning without labels}
		\For{$i=1$ \textbf{to} ${T}$}{
			\For{$\mathbf{each\ } \mathbf{X}^{(j)}\in D_{train}$ }{
				\For{$i=1$ \textbf{to} ${n}$}
				{$\text{draw two operation }\tilde{t}^{(i)}, \hat{t}^{(i)}\sim\mathcal{T}$\\
				$\hat{\mathbf{{x}}}^{(j,i)}, \tilde{\mathbf{{x}}}^{(j,i)} \leftarrow\hat{t}^{(i)}(\mathbf{x}^{(j,i)}),\tilde{t}^{(i)}(\mathbf{x}^{(j,i)})$}
		}
		$\theta^* = \arg \min\limits_{\theta} \frac{1}{2mn}\sum\limits_{j=1}^{m}\sum\limits_{i=1}^{n}l_{cl}(f_{\theta},\tilde{\mathbf{{x}}}^{(j,i)},\hat{\mathbf{{x}}}^{(j,i)},Neg(\cdot))$
	}
	\tcp{ Decoder training with labels}
	
	$	\varphi^* = \arg\min_{\varphi}{\mathcal{L}_{task}(q_\varphi,f_{\theta^*}, D_{task})}$ \\
	
	\textup{\textbf{Return\ }}
	the trained  encoder $f_{\theta^*}$ and the decoder $q_{\varphi^*}$
	\caption{The contrastive learning-based time-series foundation models on downstream tasks. }
	\label{alg:foundation_framework}
	
\end{algorithm}

\begin{figure*}[t] 
	\centering 
	\includegraphics[width=1\textwidth]{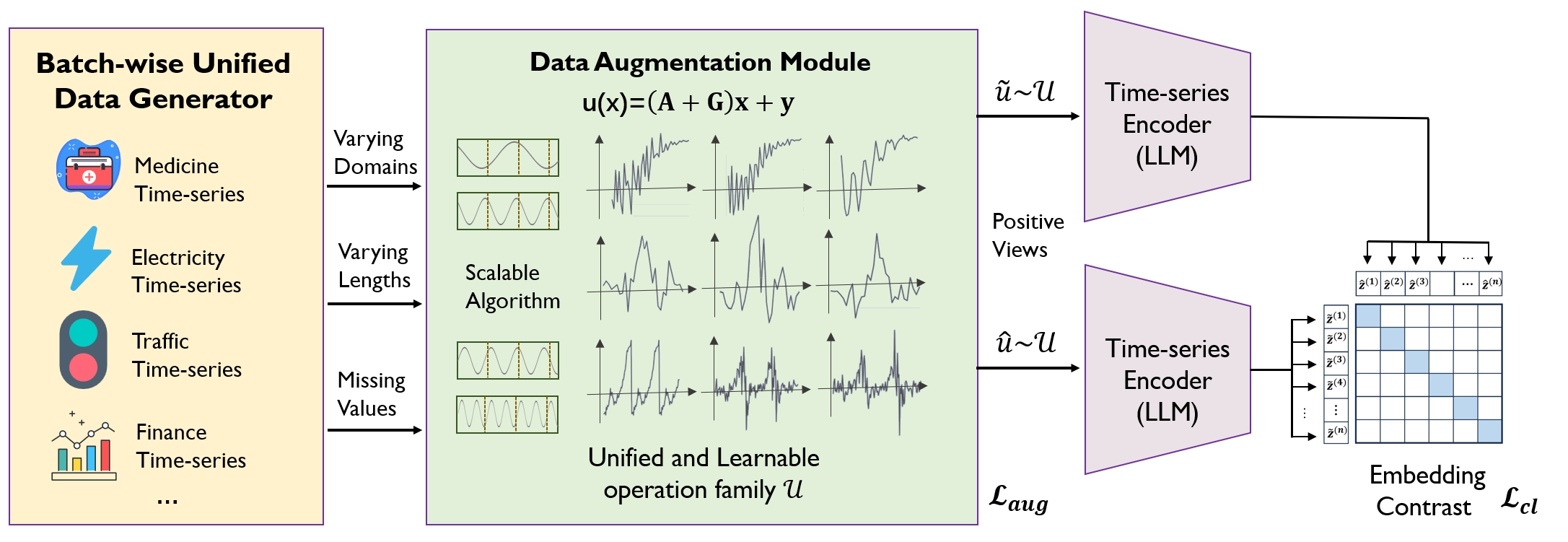} 
	\caption{Overview of the UniCL} 
	\label{fig:overview} 
\end{figure*}

Existing contrastive learning approaches mainly differ in positive view generation, and can be classified into two types: context-based and augmentation-based. Context-based approaches~\cite{yue2022ts2vec,tonekaboni2021unsupervised,kiyasseh2021clocs} generally advocate contextual consistency, considering sub-series with close temporal relationships as positive views. For example, TS2Vec~\cite{yue2022ts2vec} randomly selects two overlapping time segments as positive pairs. Other researchers~\cite{tonekaboni2021unsupervised,kiyasseh2021clocs} opt for temporal neighborhood sub-series as positive pairs. However, akin to masked modeling, context-based methods are reliant on observed data and may perform poorly on unseen data, thereby failing to address the issue of data-scarcity. Instead, augmentation-based methods can generate diverse time-series based on observed data, improving the generality of the model. Current augmentation-based methods utilize pre-defined data augmentation operation, such as jittering~\cite{yeh2023toward,ijcai2021-324,raghu2023sequential}, scaling~\cite{woo2022cost}, permutation~\cite{meng2023mhccl,poppelbaum2022contrastive}, magnitude warping~\cite{fan2020self}, masking~\cite{yue2022ts2vec,wang2024contrast}, and pooling~\cite{liang2023contrastive}. Some researchers~\cite{zhang2022self,liu2024timesurl} also apply perturbation in the frequency domain. To further improve the generality of the model, CLUDA~\cite{ozyurt2023contrastive} adopts a composition of operations to generate positive views. However, relying on pre-defined operations entails a dependence on expert knowledge~\cite{ma2023survey} and is susceptible to inductive bias~\cite{saunshi2019theoretical}. Some researchers~\cite{ijcai2021-324,yue2022ts2vec} have observed that data augmentation is data-dependent, and inappropriate augmentations can lead to poorly learned representations. A more recent study~\cite{luo2023time} employs meta-learning to select augmentation operations adaptively  based on criteria of fidelity and variety. Nonetheless, they still rely on a pre-defined set of augmentations.

	\subsubsection{Fine-tuning Time-series Foundation Model on Downstream Tasks}
	Then, the time-series foundation model $f_{\theta^\dagger}$ pre-trained in Sec.~\ref{sssec:foundation_model} can be used in time-series classification and forecasting tasks.
%
		  In general,  there are two manners to apply the pretrained foundation model for the downstream tasks~\cite{wang2024contrast}, i.e., \textit{Partial Fine-Tuning (P-FT)} and  \textit{Full Fine-Tuning (F-FT)}. The main difference between P-FT and F-FT is the parameter size of the foundation models and decoders. 
		\begin{itemize}[leftmargin=*]

	\item \textbf{Partial Fine-Tuning (P-FT).} P-FT approaches~\cite{yue2022ts2vec,wang2024contrast} are to keep the foundation model $f_{\theta^{\dagger}}$ frozen. They only train the decoder $q_\varphi$ in downstream tasks. The objective is defined as follows.
				\begin{align}\label{eq:cl_nc2}
				\varphi^* = \arg\min_{\varphi}{\mathcal{L}_{task}(q_\varphi,f_{\theta^{\dagger}}, D_{task})}
			\end{align}
		
		\item \textbf{Full Fine-Tuning (F-FT).} F-FT approaches~\cite{zhou2023onefitsall,chen2023gatgpt,chang2023llm4ts,liu2024taming} involve training both the decoder $q_\varphi$ and the foundation model $f_\theta^{\dagger}$, encompassing parameters such as positional embeddings and LayerNorm parameters~\cite{zhou2023onefitsall,chen2023gatgpt}, or the parameters within self-attention layers~\cite{chang2023llm4ts,liu2024taming} in downstream tasks, which is defined as follows.
	\begin{align}\label{eq:cl_nc1}
		\varphi^*, \theta^{*} = \arg\min_{\varphi}{\mathcal{L}_{task}(q_\varphi,f_{\theta^\dagger}, D_{task})}
	\end{align}

		\end{itemize}

Compared with P-FT,  F-FT can optimize parameters of the foundation models for downstream tasks, expecting to achieve better performance than P-FT.
Therefore,	in this paper, we use F-FT to apply our model and baselines in the downstream tasks.

\subsubsection{Variable independence.}
 Many transformer-based learning models utilize the variable-mixing (or channel-mixing) configuration~\cite{wu2021autoformer, zhou2022fedformer}, where the multivariate time-series $\mathbf{X}\in\mathbb{R}^{n\times T}$ is mapped into a timestamp-wise shared space $\mathbf{Z}\in\mathbb{R}^{T\times D}$ via an embedding layer. However, this approach introduces two critical issues: 1) The embedding layer requires the pre-definition of the number of variables, which lacks generality for cross-domain pretraining; 2) A timestamp-wise shared embedding space may not be suitable for all domains, as the mechanism of dependency can vary (e.g., lag-features of financial time-series~\cite{rasul2023lag}). To facilitate universal pretraining, our study adopts the recent widely-adopted variable independence configuration~\cite{Yuqietal-2023-PatchTST, zhou2023onefitsall,jin2023time}, processing $n$ variables independently.


\section{Framework Overview}

In this study, we introduce UniCL, a universal contrastive learning framework designed for time-series analysis. UniCL is fundamentally general and effective, capable of handling heterogeneous cross-domain time-series data with varying input lengths, based on a unified augmentation operation that generates diverse positive samples, thereby facilitating robust contrastive learning.

We provide the overview of UniCL in Fig.~\ref{fig:overview}. Primarily, there
are four steps: (1) data generation, (2) unified and scalable data augmentation module, (3) time-series encoder based on LLMs, and (4) embedding contrast. Here, we briefly clarify these steps: (1) to begin with,
the time-series datasets from diverse domains are initially partitioned into batches, shuffled, and then randomly fed into the augmentation module ;
(2) for each batch,  the proposed scalable algorithm with bounded convergence loss can deal with varying lengths of inputs with missing values, and the unified and learnable augmentation operation is employed to generate diverse and pattern-preserved positive views for contrastive learning ;
(3) the CLIP-based encoder generates embeddings for all views, effectively capturing cross-domain and general time-series patterns; 
(4) a typical contrastive loss can be employed to enhance the discriminative power of the learned embeddings.

\section{Unified Foundation Model}\label{sec:framework}
 We first demonstrate our key observations about the bias in time-series representation caused by pre-determined augmentation methods. Then, we summarize existing methods and propose a unified and learnable augmentation operation family with theoretical guarantee. To facilitate the training of such operations, we introduce two novel efficient loss functions. Additionally, we propose a scalable version of this unified operation set to handle datasets from various domains with different lengths and missing values. 
Finally, we introduce the encoder of the UniCL and the whole pre-training paradigm.

\subsection{A Unified Data Augmentation Operation}

\subsubsection{Motivational observation of bias in embedding}
\label{motivation}

As discussed in Sec.~\ref{sec:intro}, existing pre-defined time-series augmentation methods introduce an \textbf{\textit{inductive bias}} into representation learning amidst augmentation contrast. We present our key motivational observation: \textit{the bias in time-series embedding correlates positively with the \textbf{spectral distance} (SD) between raw and augmented series.} To illustrate, we first quantify the bias introduced by the pre-defined data augmentation family $\mathcal{T}(\cdot)$ as follows, in line with a previous work~\cite{zhang2022costa}. Let $\{t^{(k)}(\mathbf{x}^{(j,i)})\}_{k=1}^K$ denotes the augmentation set of $\mathbf{x}^{(j,i)}$ with size $K$, where $t^{(k)}\sim\mathcal{T}$. Let $\mathcal{T}(\mathbf{x}^{(j,i)})$ symbolize the transformation distribution of $\mathbf{x}^{(j,i)}$. Then,


\begin{equation}
	\label{bias}
	\begin{aligned}
		\text{Bias}(\mathcal{T}(\mathbf{x}^{(j,i)})) &= 
		\Big \lVert 
		\mathbb{E}_{
			t\sim\mathcal{T}
		}
		\big [f_{\theta} (t(\mathbf{x}^{(j,i)}))\big]-f_{\theta}(\mathbf{x}^{(j,i)})
		\Big \rVert_2 \\
		& \approx \Big \lVert 
		\frac{1}{K} \sum_{k=1}^{K}  
		f_{\theta} (t^{(k)}(\mathbf{x}^{(j,i)}))-f_{\theta}(\mathbf{x}^{(j,i)})
		\Big \rVert_2
	\end{aligned}
\end{equation}

\noindent Let $\mathcal{F}(\cdot)$ denote the Fast Fourier Transform (FFT), and $\left|\cdot\right|$ denote the amplitude operator, which calculates the amplitude as $\left|\cdot\right|=\sqrt{\mathcal{R}(\cdot)^2+\mathcal{J}(\cdot)^2}$, where $\mathcal{R}(\cdot)$ and $\mathcal{J}(\cdot)$ represent the real and imaginary part operators, respectively. Due to the conjugate symmetry of the frequency domain, we stipulate that the $\left|\cdot\right|$ operator only generates the first half and removes the zero-frequency component~\cite{wu2023timesnet}, \textit{i.e.,} $\left|\cdot\right|: \mathbb{C}^T\rightarrow\mathbb{R}^{\left\lfloor\frac{T}{2}\right\rfloor}$. Then, the \textit{spectral distance} between $\mathbf{x}^{(j,i)}$ and $t(\mathbf{{x}}^{(j,i)})$ can be defined as:

\begin{equation}
	\label{spectraldistance}
	SD(\mathbf{x}^{(j,i)},t(\mathbf{{x}}^{(j,i)})) = \big\lVert \small|\mathcal{F}(\mathbf{x}^{(j,i)})\small|- \small|\mathcal{F}(t(\mathbf{{x}}^{(j,i)}))\small|\big\rVert_2^2
\end{equation}

\begin{figure}[t]
	\begin{minipage}[t]{0.5\linewidth}
		\centering
		\includegraphics[width=1\textwidth]{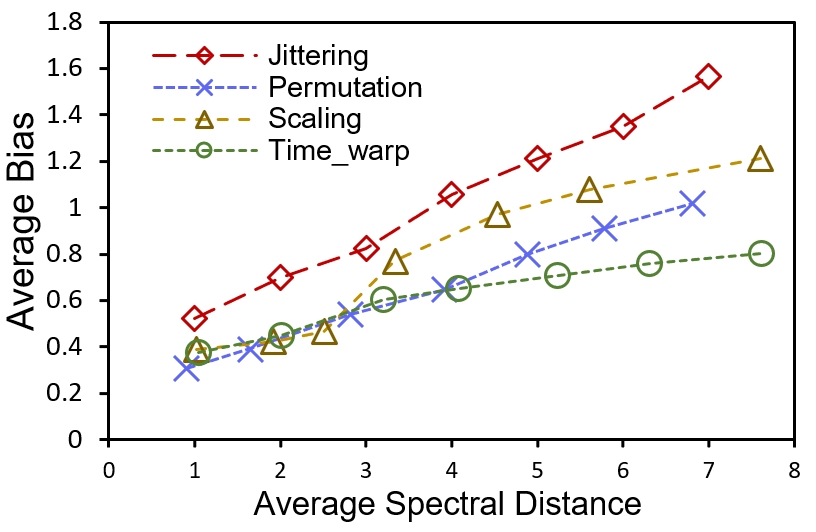}
		\captionsetup{skip=0pt,font={small,stretch=0.9}}
		\caption*{(a) Spectral Distance and Bias}
	\end{minipage}%
	\begin{minipage}[t]{0.5\linewidth}
		\centering
		\includegraphics[width=1\textwidth]{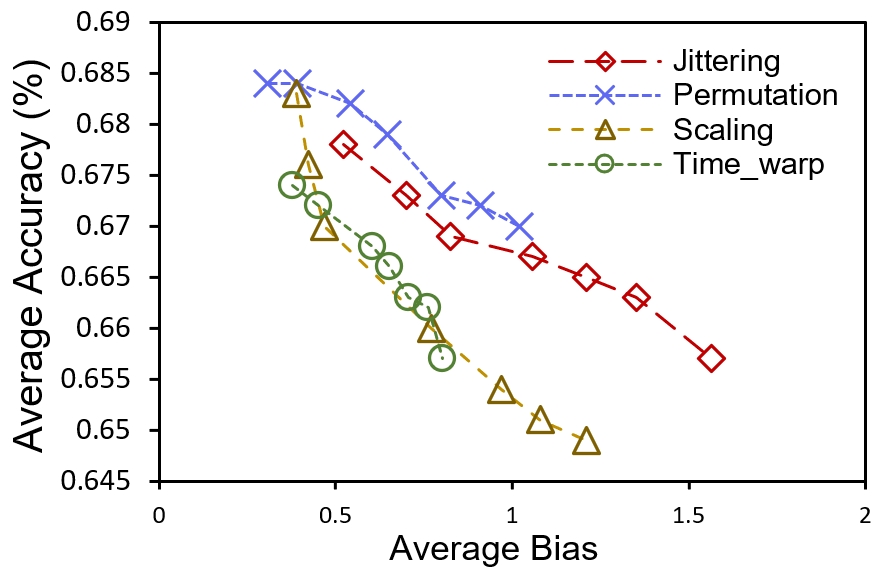}
		\captionsetup{skip=0pt,font={small,stretch=0.9}}
		\caption*{(b) Bias and accuracy}
	\end{minipage}
	\begin{minipage}[t]{0.5\linewidth}
		\centering
		\includegraphics[width=1\textwidth]{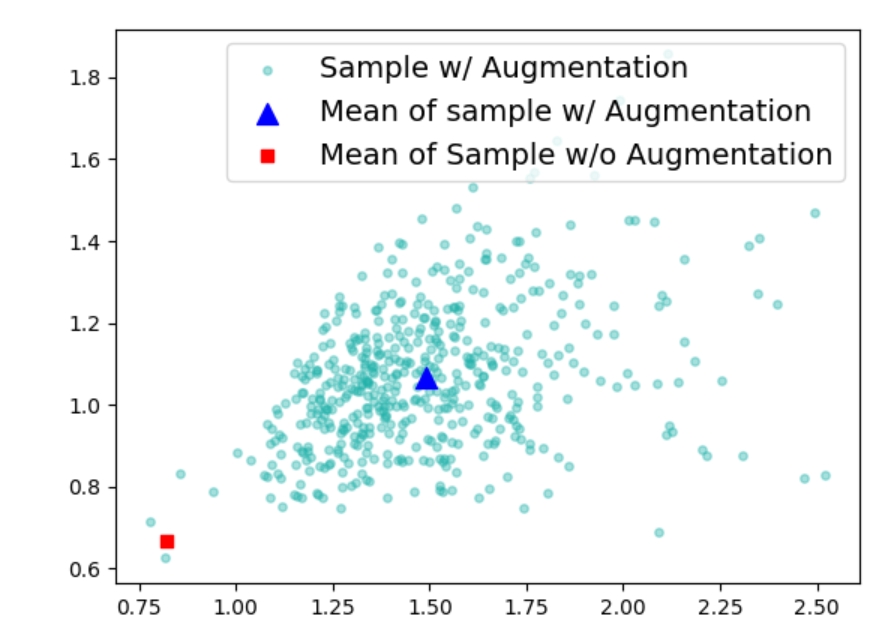}
		\captionsetup{skip=0pt,font={small,stretch=0.9}}
		\caption*{(c) High-bias embedding}
	\end{minipage}%
	\begin{minipage}[t]{0.5\linewidth}
		\centering
		\includegraphics[width=1\textwidth]{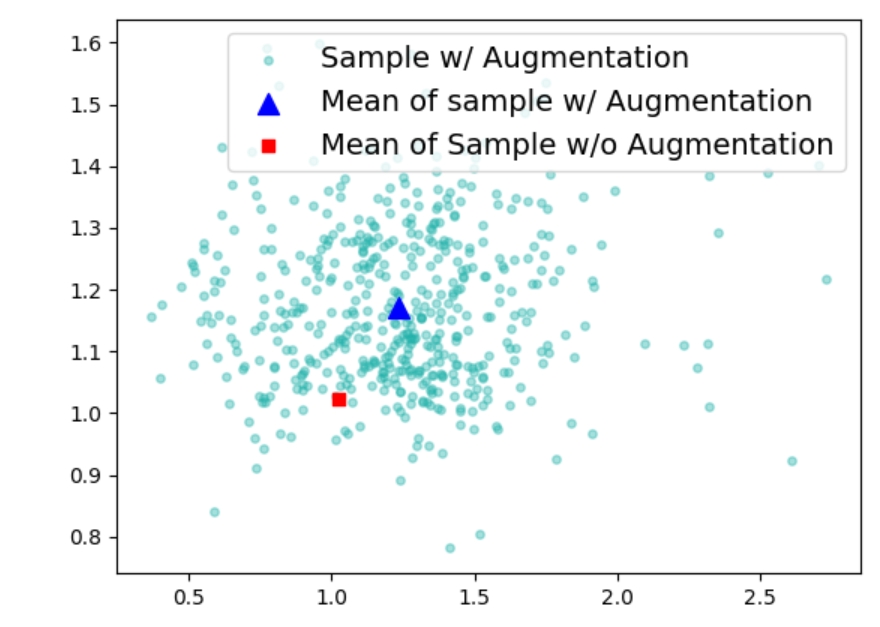}
		\captionsetup{skip=0pt,font={small,stretch=0.9}}
		\caption*{(d) Low-bias embedding}
	\end{minipage}
	\captionsetup{skip=3pt}
	\captionsetup{justification=justified,singlelinecheck=false,font={small,stretch=0.9}}
	\caption{The bias observation of four pre-defined augmentation methods and each has seven variants. Fig.1a illustrates the relationship between the average spectral distance of augmented and raw series and the average bias of learned embeddings. The y-axis represents bias, while the x-axis denotes the average spectral distance. Fig.1b demonstrates the impact of embedding bias on downstream classification task. Here, the y-axis indicates the average accuracy, and the x-axis represents bias. Fig.1c and Fig.1d visualize the high-bias and low-bias of the embedding generated by the permutation. }
	\label{motivational}
\end{figure}

We employ the representative and state-of-the-art contrastive learning method TS2Vec~\cite{yue2022ts2vec} to test 4 pre-defined augmentation methods on 23 selected multivariate datasets from the UEA Time Series Classification Archive~\cite{bagnall2018uea} and report the average bias and spectral distance. For each dataset, we train the TS2Vec encoder with the same configuration as the original paper but varying the augmentation method. These methods encompass jittering, scaling, time warping, and permutation, each offering diverse variants achieved by adjusting hyper-parameters (e.g., the standard deviation of jittering). The output layer of the encoder has two dimensions to facilitate visualization. Following, we generate $K = 500$ augmented samples randomly for each sample, and compute the average bias $\frac{1}{mn}\sum_{j=1}^{m}\sum_{i=1}^{n}\text{Bias}(\mathcal{T}(\mathbf{x}^{(j,i)}))$ and the average spectral distance $\frac{1}{mnK}\sum_{j=1}^{m}\sum_{i=1}^{n}\sum_{k=1}^{K}SD(\mathbf{x}^{(j,i)},t^{(k)}(\mathbf{x}^{(j,i)}))$, respectively. Fig.~\ref{motivational}(a) illustrates a positive correlation between the average bias and the average spectral distance of augmented and raw series. Meanwhile, Fig.~\ref{motivational}(b) demonstrates that greater bias in the embeddings results in reduced performance on downstream classification tasks. This observation motivates the need for time-series augmentation methods to control the spectral distance between augmented and raw time-series. As depicted in Fig.~\ref{motivational}(c,d), within contrastive learning, a significant bias may hinder the effectiveness of separating augmented embeddings across different instances (\textit{e.g.}, $f_{\theta}(\mathbf{\tilde{x}}^{(j,i)}), Neg(\mathbf{\tilde{x}}^{(j,i)})$), thus limiting the discriminative power of learned embeddings for downstream tasks~\cite{zhang2022costa}.

\subsubsection{\textbf{Unified Data Augmentation Operation}}
To mitigate the problem of inductive bias caused by pre-defined augmentation operation $\mathcal{T}$, one
straightforward way is to employ all augmentation operations~\cite{luo2023time, zhang2022self, ozyurt2023contrastive}. However, for some augmentation operations, the hyper-parameter space is continuous (e.g., standard deviation of jittering, number of speed changes of time warping), making it infeasible to explore the full augmented view space, especially considering their compositions~\cite{ozyurt2023contrastive}. To address this challenge, we introduce a unified operation family $\mathcal{U}$. Formally, given input time-series $\mathbf{x}^{(j,i)}$, the augmentation operation $u$ sampled from $\mathcal{U}$, \textit{i.e.}, $u\sim\mathcal{U}$, is defined as:
\begin{equation}
	 u(\mathbf{x}^{(j,i)})=\mathbf{A}\mathbf{x}^{(j,i)}+\mathbf{y}
\end{equation}
where $\mathbf{x}^{(j,i)}\in\mathbb{R}^{T}$,
$\mathbf{A}\in\mathbb{R}^{T\times T}$ and $\mathbf{y}\in\mathbb{R}^{T}$. We provide Proposition 1 to demonstrate that the operation $u\sim\mathcal{U}$ yield an augmented view space equivalent to that of each pre-defined operation and their compositions.

\begin{prop}
	Existing time-series augmentation operation set $\mathcal{T}$ includes jittering, scaling, magnitude warping, masking, pooling, and permutation. Then, the augmented view space of $\mathbf{x}^{(j,i)}$ generated by the unified operation $u\sim\mathcal{U}$ is the same as the view space generated by $t\sim\mathcal{T}$ as well as their compositions  $\bar{t} = t^{(n)} \circ t^{(n-1)}\circ\cdots\circ t^{(1)}$, where $t^{(i)}\sim\mathcal{T}$.
\end{prop}

\begin{proof}
	The proof for each augmentation operation is presented in Table~\ref{proof}. Therefore, for each $t^{(i)}\sim\mathcal{T}$, there exist matrix $\mathbf{A}^{(i)}$ and vector $\mathbf{y}^{(i)}$ such that $t^{(i)}(\mathbf{x}^{(j,i)})=\mathbf{A}^{(i)}\mathbf{x}^{(j,i)}+\mathbf{y}^{(i)}\sim\mathcal{U}$. Let $\bar{t} = t^{(n)} \circ t^{(n-1)}\circ\cdots\circ t^{(1)}$ be the composite operator. Then,
	\vspace{0cm}
	\begin{equation}
		\begin{aligned}
			\bar{t}(\mathbf{{x}}) &= \mathbf{A}^{(n)}\cdot\bigg(
			\mathbf{A}^{(n-1)}\cdot\big(
			\cdots \small(
			\mathbf{A}^{(1)}\mathbf{x}^{(j,i)}+\mathbf{y}^{(1)}
			\small)\dots
			\big) + \mathbf{y}^{(n-1)}
			\bigg) + \mathbf{y}^{(n)}\\
			&= \mathbf{A}^{(n)}\cdot \mathbf{A}^{(n-1)}\cdots\mathbf{A}^{(1)}\mathbf{x}^{(j,i)} + \sum_{i=2}^{n}\mathbf{A}^{(i)}\mathbf{y}^{(i-1)} + \mathbf{y}^{(n)}\\
			&= \bar{\mathbf{A}}\mathbf{x}^{(j,i)}+\bar{\mathbf{y}}
		\end{aligned}
	\end{equation}
	\vspace{-0.2cm}

\noindent The above concludes that the compositions of $t$ also belongs to the unified operation family $\mathcal{U}$.
\end{proof}
	\vspace{-0.1cm}
	\begin{table}[h]

		\setlength{\abovecaptionskip}{0.2cm} 
		\setlength{\belowcaptionskip}{-0.2cm}
		\caption{Proof of unified operation}
				\centering
		\label{notations}
		\begin{tabular}{lll}
			\toprule 
	
			Name & $t(\mathbf{x}^{(j,i)})$ & $u(\mathbf{x}^{(j,i)})=\mathbf{A}\mathbf{x}^{(j,i)}+\mathbf{y}$\\ 
			\midrule 
			Jittering & 
			\makecell[l]{$\{x^{(j,i)}_k+\epsilon_k\}_{k=1}^{T}$,\\ $\epsilon_k\sim\mathcal{N}(\mu,\sigma)$} & 
			$\mathbf{A}=\mathbf{I}$, $\mathbf{y} = (\epsilon_1,\ldots, \epsilon_{T})^T$\\
			\hline
			\specialrule{0cm}{1pt}{1pt}
			Scaling &
			$\{ax^{(j,i)}_k\}_{k=1}^{T},a\in\mathbb{R}$ &
			$\mathbf{A}=Diag\{a,\cdots,a\}$,
			$\mathbf{y}=\mathbf{0}$\\
			\hline
			\makecell[l]{Magnitude \\warping} & 
			$\{a_kx^{(j,i)}_k\}_{k=1}^{T},a_k\in\mathbb{R}$ &
			\makecell[l]{$\mathbf{A}=Diag\{a_1,\cdots,a_{T}\}$,\\
			$\mathbf{y}=\mathbf{0}$}\\
			\hline
			\specialrule{0cm}{1pt}{2pt}
			Masking &
			\makecell[l]{$\{a_kx^{(j,i)}_k\}_{k=1}^{T}$,\\
				$a_k=\{0,1\}$} &
			\makecell[l]{$\mathbf{A}=Diag\{a_1,\cdots,a_{T}\}$,\\
			$\mathbf{y}=\mathbf{0}$}\\
			\hline
			
			\specialrule{0cm}{1pt}{1pt}
			\makecell[l]{Mean \\ Pooling} &
			\makecell[l]{The $k$-th bin=$\{\frac{1}{w}\cdot$\\
				$\sum_{h=w(k-1)+1}^{wk}x^{(j,i)}_h\}^{w}$
		}&
			\makecell[l]{$\mathbf{A}=\left (\begin{smallmatrix}
				\mathbf{B} &  & \mathbf{0} \\
				 & \ddots &  \\
				 \mathbf{0} &  & \mathbf{B} \\
			\end{smallmatrix}\right )$, $\mathbf{B}=\frac{1}{w}\cdot\mathbf{1}_w$\\
			$\mathbf{y}=\mathbf{0}$
		}\\
		\hline
		Permutation &
		\makecell[l]{$\{x^{(j,i)}_{\pi(k)}\}_{k=1}^{T}$,\\
		$\pi\in\mathcal{S}_T$ } &
		\makecell[l]{$\mathbf{A}=\left (\begin{smallmatrix}
				\mathbf{e}_{\pi(1)}^T \\
				 \cdots   \\
				\mathbf{e}_{\pi(T)}^T \\
			\end{smallmatrix}\right )$, 	$\mathbf{y}=\mathbf{0}$
		}\\
			\bottomrule 
		\end{tabular}
		\label{proof}
	\end{table}
\vspace{-0.3cm}


To introduce randomness into the generation of diverse positive samples, without loss of generality, we can incorporate a random matrix $\mathbf{G}$ with the deterministic matrix $\mathbf{A}$. Formally, given input time-series with length $T$, the \textbf{non-scalable unified operation} can be expressed as:

\begin{equation}
	\label{augmentation}
	u(\mathbf{x}^{(j,i)})=(\mathbf{A}+\mathbf{G})\mathbf{x}^{(j,i)}+\mathbf{y}
\end{equation} 
and the matrix form is:

\begin{equation}
	\label{matrix_form}
	u(\mathbf{X}^{(j)})=\mathbf{X}^{(j)}(\mathbf{A}+\mathbf{G})^{T}+\mathbf{y}
\end{equation} 

where $\mathbf{A}$ is a $T\times T$ deterministic matrix, $\mathbf{G}$ is a $T\times T$ random matrix, and $\mathbf{y}$ is a $T$ dimensional random vector. 
As shown in Alg.~\ref{alg:non_scalable}, we set $\mathbf{G}$ to be a Gaussian noise matrix (line 2), where all elements are independent and identically distributed (\textit{i.i.d.}) random variable following Gaussian distribution, \textit{i.e.}, $\mathbf{G}[i][j]\sim\boldsymbol{N}\big(\boldsymbol{\mu}^{(G)}[i][j],(\boldsymbol{\sigma}^{(G)}[i][j])^2\big)$. Here, both $\boldsymbol{\mu}^{(G)}$ and $\boldsymbol{\sigma}^{(G)}$ are $T\times T$ trainable matrix. Similarly, $\mathbf{y}$ is a Gaussian noise vector (line 3), where all elements are \textit{i.i.d.} and each $y_i$ follows a Gaussian distribution $y_i\sim\boldsymbol{N}\big(\mu_{i}^{(y)},(\sigma_{i}^{(y)})^2\big)$ with random vectors $\boldsymbol{\mu}^{(y)}$ and $\boldsymbol{\sigma}^{(y)}$ as trainable parameters. Therefore, the time and space complexity is $O(4T^2+3T)$, and we will introduce scalable and efficient algorithms in Section~\ref{sec:scal}. Let $\mathcal{U}(\mathbf{x}^{(j,i)})$ symbolize the transformation distribution of $\mathbf{x}^{(j,i)}$, we have the following proposition:
\begin{prop}
	\label{distribution}
	The transformation distribution $\mathcal{U}(\mathbf{x}^{(j,i)})$ follows a multivariate normal distribution.
\end{prop}
\begin{proof}
	 Considering the $k$-th random variable $(\mathcal{U}(\mathbf{x}^{(j,i)}))_k$, we have:
	\begin{equation}
		\begin{aligned}
		\mathcal{U}(\mathbf{x}^{(j,i)})_k &= (\mathbf{A}\mathbf{x}^{(j,i)})_k+ (\mathbf{G}\mathbf{x}^{(j,i)})_k + y_k\\
		&= (\mathbf{A}\mathbf{x}^{(j,i)})_k +
		\sum_{h=1}^{T}\mathbf{G}[k][h]x^{(j,i)}_h+ y_k
		\end{aligned}
	\end{equation} 
	Notice that the first term is constant, as the matrix $\mathbf{A}$ and input series $\mathbf{x}^{(j,i)}$ are deterministic. The remainder of the equation becomes a linear combination of normal distributions and \textit{i.i.d.} random variables, \textit{i.e.,} $\mathbf{G}[i][j]$ and $y_k$, resulting in $(\mathcal{U}(\mathbf{x}^{(j,i)}))_k$ following normal distribution. The above concludes that the transformation distribution
	$\mathcal{U}(\mathbf{{x}})$ is a multivariate normal distribution.
\end{proof}

\begin{algorithm}[t]
	\KwIn{Time-series $\mathbf{x}^{(j,i)}\in\mathbb{R}^{T}$, $T\times T$ Matrices $\mathbf{A}, \boldsymbol{\mu}^{(G)}, \boldsymbol{\sigma}^{(G)}$, $T$-dimension Vectors $\boldsymbol{\mu}^{(y)}, \boldsymbol{\sigma}^{(y)}$.}
	\KwOut{ The augmented time-series}
\tcp{Generate Gaussian noise}
$\mathbf{G}, \mathbf{y}\leftarrow rand(-1,1)$\\
$\mathbf{G}\leftarrow \mathbf{G}\odot\boldsymbol{\sigma}^{(G)}+\boldsymbol{\mu}^{(G)}$\\
$\mathbf{y}\leftarrow \mathbf{y}\odot\boldsymbol{\sigma}^{(y)}+\boldsymbol{\mu}^{(y)}$\\
\textup{\textbf{Return\ }}
$(\mathbf{A}+\mathbf{G})\mathbf{x}^{(j,i)}+\mathbf{y}$
\caption{Non-scalable algorithm of unified operation}
\label{alg:non_scalable}

\end{algorithm}

\subsection{Scalable  and Diverse Data Augmentation}
\label{sec:scal}

    We first propose  a data augmentation objective based on our proposed unified operation to generate spectrum-preserved and diverse time-series data. Then, we propose a scalable algorithm to apply this objective to time-series data with various lengths. 


\subsubsection{{A Spectrum-preserved and  Diverse Objective}}
\label{sec:objectives}
In contrastive learning, augmented data should satisfy two properties to be effective: pattern preservation and diversity~\cite{luo2023time}. Augmented data should preserve the essential pattern of the original data, ensuring that the similarity relationships between instances are maintained. Then, the $f_{\theta}$ optimized by maximizing the similarity
of positive views are expected to learn representations that capture time-series intrinsic patterns. Additionally, augmented data should introduce diversity into generated time-series, enabling the model to learn robust representations that generalize well to unseen data. We introduce two novel losses to facilitate the learning of our augmentation module: spectrum-preservation loss and spectrum-diversity loss.

\textit{1) \textbf{Spectrum-preservation loss.} } As discussed in Sec.~\ref{motivation}, in order to generate low-bias embeddings, the positive pairs $\tilde{\mathbf{x}}^{(j,i)}$ and $\hat{\mathbf{x}}^{(j,i)}$ should be close to the original series $\mathbf{{x}}$ in terms of spectral distance. Therefore, the spectral distance in Eq.~\ref{spectraldistance} can serve as the metric to measure the pattern differences of augmented series and raw series. Formally, the spectrum-preservation loss $l_{p}$ can be defined as:
\begin{equation}
	\label{eq:preservation_loss}
	\begin{aligned}
	l_p(\mathcal{U},\mathbf{x}^{(j,i)}) = \frac{1}{2}\mathbb{E}_{
		\tilde{u}\sim\mathcal{U}
	}&
	\big[
	\big\lVert
	|\mathcal{F}\small(\tilde{u}\small(\mathbf{x}^{(j,i)}\small))|-
	|\mathcal{F}\small(\mathbf{x}^{(j,i)}\small)|
	\big\rVert_2^2
	\big] \\	 
	+\frac{1}{2} & \mathbb{E}_{
		\hat{u}\sim\mathcal{U}}
		\big[
	\big\lVert
	|\mathcal{F}\small(\hat{u}\small(\mathbf{x}^{(j,i)}\small))|-
	|\mathcal{F}\small(\mathbf{x}^{(j,i)}\small)|
	\big\rVert_2^2
	\big] \\
	= \mathbb{E}_{
		{u}\sim\mathcal{U}
	}\big[
	\big\lVert&
	|\mathcal{F}\small(u\small(\mathbf{x}^{(j,i)}\small))|-
	|\mathcal{F}\small(\mathbf{x}^{(j,i)}\small)|
	\big\rVert_2^2
	\big] \\
	\end{aligned}
\end{equation}
\textit{2) \textbf{Spectrum-diversity loss.} } To enhance the diversity of the positive pairs $\tilde{\mathbf{x}}^{(j,i)}$ and $\hat{\mathbf{x}}^{(j,i)}$, we need to: \textbf{1)} define a metric to quantify the diversity of positive pairs, and \textbf{2)} identify which patterns in $\mathbf{x}^{(j,i)}$ are not essential and can therefore be diversified. 

By Prop.~\ref{distribution}, the positive pairs $\tilde{\mathbf{x}}^{(j,i)}$ and $\hat{\mathbf{x}}^{(j,i)}$ are two random vectors, where the $k$-th values $\tilde{x}^{(j,i)}_k$ and $\hat{x}^{(j,i)}_k$ are random variables with \textit{i.i.d.} normal distribution. One intuitive approach to measure the diversity is to compute the average entropy of all random variables, given by $\frac{1}{T}\sum_{k=1}^{T}\ln(2\pi e\sigma_k^2)$, where $\sigma_k^2=\sum_{h=1}^{T}(\boldsymbol{\sigma}^{(G)}[k][h])^2$ $x^{(j,i)}_h+(\sigma_k^{(y)})^2$. However, simply increasing the entropy of each point results in large $\boldsymbol{\sigma}^{(G)}$ and $\boldsymbol{\sigma}^{(y)}$, introducing meaningless noise. 
More generally, suppose the distribution of $\tilde{x}^{(j,i)}_k$ and $\hat{x}^{(j,i)}_k$ are unknown. Another intuitive approach to measure the diversity of positive views is to compute the average Kullback-Leibler ($KL$) divergence between the probability distributions of each pair of random variables at the same time point, given by $\frac{1}{T}\sum_{k=1}^{T}KL(P(\tilde{x}^{(j,i)}_k||$ $P(\hat{x}^{(j,i)}_k))$, where $P(\cdot)$ denotes the probability density function (PDF). However, in a time-series, only one observation is available at each timestamp. Estimating $P(\tilde{\mathbf{x}}^{(j,i)}_k)$ and $P(\hat{\mathbf{x}}^{(j,i)}_k)$ requires sampling $2M$ operators $u\sim\mathcal{U}$, with $M$ denoting the number of samplings for each view. Since this approximation must be conducted for every time point, the time complexity amounts to $O(2M\cdot T)$, which becomes infeasible when both $M$ and $T$ are large. 

To tackle the complexity issue, rather than approximating the distribution in the time domain, we transform the positive views $\tilde{\mathbf{{x}}}^{(j,i)}$ and $\hat{\mathbf{{x}}}^{(j,i)}$ into the frequency domain as $|\mathcal{F}(\tilde{\mathbf{{x}}}^{(j,i)})|$ and $|\mathcal{F}(\hat{\mathbf{{x}}}^{(j,i)})|$, respectively. 
Here, the $k$-th element $|\mathcal{F}(\tilde{\mathbf{{x}}}^{(j,i)})|_k$ and $|\mathcal{F}(\hat{\mathbf{{x}}}^{(j,i)})|_k$, denoting the amplitude of the $k$-th frequency component, are also random variables. We convert the  amplitude sequence to a probability mass function (PMF) by $P(\tilde{\mathbf{{x}}}^{(j,i)})=Softmax(|\mathcal{F}(\tilde{\mathbf{{x}}}^{(j,i)})|/\tau)$ and $P(\hat{\mathbf{{x}}}^{(j,i)})=Softmax(|\mathcal{F}(\hat{\mathbf{{x}}}^{(j,i)})|/\tau)$, where $\tau$ is the temperature parameter. Then, we can measure the diversity by calculating the Jensen-Shannon ($JS$) divergence between PMF $P(\tilde{\mathbf{{x}}}^{(j,i)})$ and PMF $P(\hat{\mathbf{{x}}}^{(j,i)})$, which is more efficient.

However, not all the frequency component should be diversified. 
Given $P(\tilde{\mathbf{{x}}}^{(j,i)})=[\tilde{p}_1,\tilde{p}_2,\cdots,\tilde{p}_{\left\lfloor\frac{T}{2}\right\rfloor}]$, the $\tilde{p}_k$ represent the relative strength of $k$-th frequency within the entire spectrum. The low-frequency component ($k\rightarrow1$) carries crucial semantic information such as trend and periodicity, while high-frequency component ($k\rightarrow \left\lfloor\frac{T}{2}\right\rfloor$) usually carries meaningless noise~\cite{zhou2022fedformer}. Therefore, we multiply the PMF by a decay factor $\boldsymbol{\alpha}=[\alpha_1,\alpha_2,\cdots,\alpha_{\left\lfloor\frac{T}{2}\right\rfloor}]^T$ using element-wise multiplication $\odot$, thereby assigning different weights to different components. Formally, the spectrum-diversity loss $l_{d}$ can be defined as:
\begin{equation}
	\label{eq:diversity_loss}
\begin{aligned}
	&l_d(\mathcal{U},\mathbf{x}^{(j,i)},\boldsymbol{\alpha}) =\\
	&\mathbb{E}_{
		(\tilde{u},\hat{u})\sim\mathcal{U}
	}
	\big[-\log
	JS\big(
	\boldsymbol{\alpha}\odot P(\tilde{u}(\mathbf{x}^{(j,i)}))
	||
	\boldsymbol{\alpha}\odot P(\hat{u}(\mathbf{x}^{(j,i)}))
	\big)
	\big]
\end{aligned}
\end{equation}
\noindent where $P(\cdot)=Softmax(\left|\mathcal{F}(\cdot)\right|/\tau)$, and $\log$ is used to stabilize the optimization.

\textit{3) \textbf{Summary.} } Formally, we define the  objective function $\mathcal{L}_{aug}$ of the data augmentation module as follows:
\begin{equation}
	\label{eq:summary_loss}
	\mathcal{L}_{aug}(\mathcal{U}, \{\mathbf{X}^{(j)}\}_{j=1}^{m}, \boldsymbol{\alpha}) = \sum_{j=1}^{m}\sum_{i=1}^{n}l_p(\mathcal{U},\mathbf{x}^{(j,i)})
	+ \lambda\cdot
	l_d(\mathcal{U},\mathbf{x}^{(j,i)},\boldsymbol{\alpha})\\
\end{equation}
where $\lambda$ is a hyper-parameter.

%
%

\subsubsection{\textbf{Scalable operations}}
\label{sec:scalable}
With the learnable unified operation and the corresponding loss function, the question is how we can efficiently employ it in time-series datasets exhibiting high variations, including varying sequence lengths and potential missing values.

\begin{figure}[t] 
	\centering 
	\includegraphics[width=0.5\textwidth]{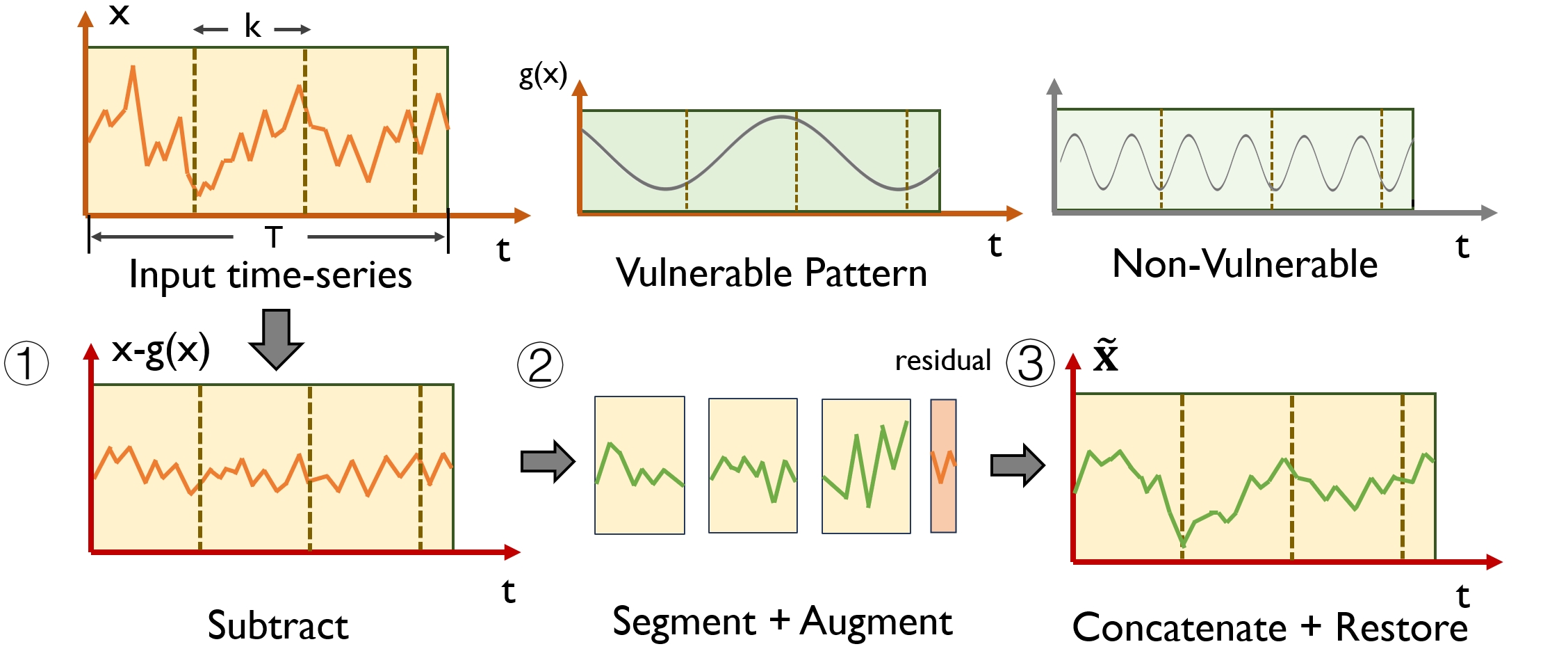} 
	\caption{The procedure of the scalable algorithm involves augmenting time-series data with varying lengths through pattern-preserved segmentation and concatenation.} 
	\label{fig:scalable} 
\end{figure}

\textit{1) \textbf{Varying sequence lengths}}. To propose a scalable algorithm for handling inputs with varying lengths, two key issues need to be addressed: 1) Given the input series $\mathbf{x}^{(j,i)}\in\mathbb{R}^{T}$, the space complexity of the non-scalable algorithm outlined in Alg.~\ref{alg:non_scalable} is $O(4\times T^2+3T)$. This complexity becomes impractical when we receive the long sequence. 2) In large-scale pretraining, different datasets contain varying lengths of instances, and employing different size of unified operation for each dataset is inefficient. Thus, we introduce a scalable algorithm that offers efficient implementation, necessitating only a fixed-size unified operation across all datasets. This approach results in a space complexity of $O(K^2)$, where $K$ is a constant. 

Particularly, with a constant $K$, we define the \textbf{fix-sized unified operation} as follows:

\begin{equation}
	\label{eq:fix_size}
	u(\mathbf{x}^{(j,i)},k)=(\mathbf{A}_{k\times k}+\mathbf{G}_{k\times k})\mathbf{x}_{k\times 1}^{(j,i)}+\mathbf{y}_{k\times 1}
\end{equation} 

 Such an operation can only handle input time-series with a fix length $K$, therefore we need to extend (resp. segment) the inputs $\mathbf{x}^{(j,i)}$ when $T<K$ (resp. $T>K$). When $T<K$, we employ iterative extension of the time series through repetition until its length equals $K$. This iterative repetition aligns with the periodicity assumption of Fourier analysis, ensuring $x_{i}=x_{i+T}$, thereby preserving the amplitude spectrum of $\mathbf{x}$. Conversely, alternative methods such as padding~\cite{zerveas2021transformer} may disrupt the spectrum pattern of $\mathbf{x}$. 
  
  As illustrated in Fig.~\ref{fig:scalable} and Alg.~\ref{alg:scalable}, when $T>K$, we employ segmentation to the input time-series. Since segmentation may disrupt the intrinsic pattern, we first denote the $g(\mathbf{x})$ as the vulnerable pattern which will be disrupted by segmentation. This pattern should be subtracted prior to the segmentation (line 1) and restored after concatenation (line 6), which is spectrum-preservation due to the linearity of the Fourier transform. 
  Then, we segment the time-series into $\lfloor\frac{T}{K}\rfloor + 1$ non-overlapping and contiguous subseries $\{\mathbf{h}^{(l)}\in\mathbb{R}^{K}, \mathbf{h}^{(\lfloor\frac{T}{K}\rfloor+1)}|l=1,\cdots,\lfloor\frac{T}{K}\rfloor\}$ (line 2), where $\mathbf{h}^{(\lfloor\frac{T}{K}\rfloor+1)}\in\mathbb{R}^{res}$ is the residual term and $ 0\leq res<K$. Subsequently, we augment each subseries separately using fix-sized unified operation (line 4) and concatenate them in the same order as $\tilde{\mathbf{x}}^{(j,i)}=[\tilde{u}^{(1)}(\mathbf{h}^{(1)},k), \cdots,\tilde{u}^{(\lfloor\frac{T}{K}\rfloor)}(\tilde{\mathbf{h}}^{(\lfloor\frac{T}{K}\rfloor)},k), \mathbf{h}^{(\lfloor\frac{T}{K}\rfloor+1)}]$, where $\tilde{u}\sim\mathcal{U}$.  Finally, we obtain the augmented series by restoration (line 6). 
  
  \textbf{Time and Space Complexity.} For line 1-3 and 5-6, the time and space complexity is $O(4T)$. For line 4, since only a fix-sized unified operation family is needed, the space complexity is $O(4K^2 +3K)$ according to Alg.~\ref{alg:non_scalable} and the time complexity is $O(\lfloor\frac{T}{K}\rfloor\cdot(4K^2 +3K))$.

 \begin{algorithm}[t]
	\KwIn{Time-series $\mathbf{x}^{(j,i)}\in\mathbb{R}^{T}$, Window size $K$, Linear function $g(\cdot)$, fix-sized unified operation $u\sim\mathcal{U}$.}
	\KwOut{ The augmented time-series}
	$\mathbf{x}^{(j,i)}\leftarrow \mathbf{x}^{(j,i)}-g(\mathbf{x}^{(j,i)})$\\
	$\mathbf{h}^{(1)}, \mathbf{h}^{(2)}\cdots, \mathbf{h}^{(\lfloor\frac{T}{K}\rfloor)}\leftarrow \mathbf{x}^{(j,i)}_{[1:K]},\mathbf{x}^{(j,i)}_{[K+1:2K]}, \cdots, \mathbf{x}^{(j,i)}_{[(\lfloor\frac{T}{K}\rfloor-1)\cdot K+1:\lfloor\frac{T}{K}\rfloor \cdot K]}$\\
	$\mathbf{h}^{(\lfloor\frac{T}{K}\rfloor+1)} \leftarrow \mathbf{x}^{(j,i)}_{[\lfloor\frac{T}{K}\rfloor\cdot K+1:]}$\\
 	$\{\tilde{\mathbf{h}}^{(1)},\cdots,\tilde{\mathbf{h}}^{(\lfloor\frac{T}{K}\rfloor)}\}\leftarrow$ Augment $\{\mathbf{h}^{(1)},\cdots,\mathbf{h}^{(\lfloor\frac{T}{K}\rfloor)}\}$ separately using $\tilde{u}\sim\mathcal{U}$ in  Eq.~\ref{eq:fix_size}.\\
 	$\tilde{\mathbf{x}}^{(j,i)}\leftarrow$ concatenate $[\tilde{\mathbf{h}}^{(1)},\cdots,\tilde{\mathbf{h}}^{(\lfloor\frac{T}{K}\rfloor)},\mathbf{h}^{(\lfloor\frac{T}{K}\rfloor+1)}]]$\\
 	$\tilde{\mathbf{x}}^{(j,i)}\leftarrow\tilde{\mathbf{x}}^{(j,i)}+g(\mathbf{x}^{(j,i)})$\\
 	\textup{\textbf{Return\ }}
 	$\tilde{\mathbf{x}}^{(j,i)}$
 	\caption{Scalable algorithm of unified operation}
 	\label{alg:scalable}
 \end{algorithm}


\begin{prop}
	\label{scalable}
	Given that \textbf{1)} $K < T$; \textbf{2)} $g(\mathbf{x}^{(j,i)})$ is a linear function (\textit{i.e.}, a linear map $g: \mathbb{R}^{T}\mapsto \mathbb{R}^{T}$): then the augmented view space generated by the scalable algorithm Alg.~\ref{alg:scalable} is a subspace of the view space generated by the non-scalable algorithm Alg.~\ref{alg:non_scalable}.
\end{prop}

\begin{proof}
We can express the equivalent form of the scalable algorithm by introducing a \textbf{scalable unified operation} family $\mathcal{S}$. For each scalable unified operation $s\sim\mathcal{S}$, we have:
	
	\begin{equation}
		\label{sclable algorithm}
		\begin{aligned}
		s(\mathbf{x}^{(j,i)},k)=(\mathbf{A}'_{T\times T}+\mathbf{G}'_{T\times T})\cdot(\mathbf{x}^{(j,i)}-g(\mathbf{x}^{(j,i)}))+\mathbf{y}'_{T\times 1} + g(\mathbf{x}^{(j,i)})
		\end{aligned}
	\end{equation}
	
 where $\mathbf{A}'_{T\times T}=Diag\{\mathbf{A}_{k\times k}^{(1)},\cdots,\mathbf{A}_{k\times k}^{(\lfloor\frac{T}{K}\rfloor)},\mathbf{I}_{res\times res} \}$ and  $\mathbf{G}'_{T\times T}= Diag\{\mathbf{G}_{k\times k}^{(1)},$ $\cdots,\mathbf{G}_{k\times k}^{(\lfloor\frac{T}{K}\rfloor)}, \mathbf{I}_{res\times res} \}$ are both block diagonal matrices, 
 and $\mathbf{y}'_{T\times 1}=[(\mathbf{y}_{k\times 1}^{(1)})^T\cdots,\mathbf{1}_{res\times1}^T]^T$. Since $g(\mathbf{x}^{(j,i)})$ is a linear function, we have $g(\mathbf{x}^{(j,i)})=\mathbf{H}\mathbf{x}^{(j,i)}$, where $\mathbf{H}$ is a $T\times T$ matrix. Thus, we get the following Equation:
 	\begin{equation}
 	\begin{aligned}
 		s(\mathbf{x}^{(j,i)},k)&=(\mathbf{A}'_{T\times T}+\mathbf{G}'_{T\times T})\cdot(\mathbf{x}^{(j,i)}-\mathbf{H}\mathbf{x}^{(j,i)})+\mathbf{y}'_{T\times 1} + \mathbf{H}\mathbf{x}^{(j,i)}\\
	&=
		\left(\left(\mathbf{A}'_{T\times T}\left(\mathbf{I-H}\right)+\mathbf{H}\right)+	\left(\mathbf{G}'_{T\times T}\left(\mathbf{I-H}\right)\right)\right)
		\mathbf{x}^{(j,i)}  + \mathbf{y}'_{T\times 1} \\
	&= (\mathbf{A}^{new} + \mathbf{G}^{new})\mathbf{x}^{(j,i)} + \mathbf{y}^{new} \sim\mathcal{U}\\
 	\end{aligned}
 \end{equation}
 
 Thus, we have $\mathcal{S}\subset\mathcal{U}$.
\end{proof}

\begin{lemma}
	\label{boundedloss}
	The difference in loss convergence between scalable and non-scalable algorithms is bounded by: \textbf{(1)} hyperparameter $\lambda$; \textbf{(2)} the spectral distance $SD(s(\mathbf{x},k), u(\mathbf{x},T))$, where $s\sim\mathcal{S},u\sim\mathcal{U}$.
\end{lemma}

\begin{proof}
	By Prop.~\ref{scalable}, the non-scalable operation $u\sim\mathcal{U}$ is associated with a larger hypothesis space compared to the scalable operation $s\sim\mathcal{S}$. Without loss of generality, we make the assumption that $	\mathcal{L}_{aug}(\mathcal{U}^*,\{\mathbf{X}^{(j)}\}_{j=1}^{m},\boldsymbol{\alpha})<	\mathcal{L}_{aug}(\mathcal{S}^*,\{\mathbf{X}^{(j)}\}_{j=1}^{m},\boldsymbol{\alpha})$, where $\mathcal{U}^*=\arg\min\limits_{\mathcal{U}}\mathcal{L}_{aug}(\mathcal{U},\{\mathbf{X}^{(j)}\}_{j=1}^{m},\boldsymbol{\alpha})$ and $\mathcal{S}^*=\arg\min\limits_{\mathcal{S}}\mathcal{L}_{aug}(\mathcal{S},$ $\{\mathbf{X}^{(j)}\}_{j=1}^{m},\boldsymbol{\alpha})$. We derive the upper bound of the difference of the convergence loss:

	\begin{small}
	\begin{equation}
	\begin{aligned}
		|\mathcal{L}_{aug}(\mathcal{S}^*,\{\mathbf{X}^{(j)}\}_{j=1}^{m},\boldsymbol{\alpha})&-\mathcal{L}_{aug}(\mathcal{U}^*,\{\mathbf{X}^{(j)}\}_{j=1}^{m},\boldsymbol{\alpha})|\\
		=  \mathcal{L}_{aug}(\mathcal{S}^*,\{\mathbf{X}&^{(j)}\}_{j=1}^{m},\boldsymbol{\alpha})-\mathcal{L}_{aug}(\mathcal{U}^*,\{\mathbf{X}^{(j)}\}_{j=1}^{m},\boldsymbol{\alpha})\\
	=
		\sum_{j=1}^{m}\sum_{i=1}^{n}l_{p}(\mathcal{S}^*&,\mathbf{x}^{(j,i)})-l_{p}(\mathcal{U}^*,\mathbf{x}^{(j,i)})+\lambda l_{d}(\mathcal{S}^*,\mathbf{x}^{(j,i)},\boldsymbol{\alpha})
		-\\
		&\lambda l_{d}(\mathcal{U}^*,\mathbf{x}^{(j,i)},\boldsymbol{\alpha})\\
		\leq
		\sum_{j=1}^{m}\sum_{i=1}^{n}l_{p}(\mathcal{S}^*&,\mathbf{x}^{(j,i)})-l_{p}(\mathcal{U}^*,\mathbf{x}^{(j,i)})+\lambda\cdot\ln2\\
		=
		\sum_{j=1}^{m}\sum_{i=1}^{n}\mathbb{E}_{s\sim\mathcal{S}^*}&
		[\big\lVert
		|F(s(\mathbf{x}^{(j,i)}))| - |F(\mathbf{x}^{(j,i)})| \big\rVert_2^2]-
		\\
		&\mathbb{E}_{u\sim\mathcal{U}^*}
		[\big\lVert
		|F(u(\mathbf{x}^{(j,i)}))|
		 - |F(\mathbf{x}^{(j,i)})| \big\rVert_2^2]+\lambda\cdot\ln2
		\end{aligned}
	\end{equation}
	\end{small}

	Therefore, to lower the upper bound, we can 1) choose small $\lambda$; 2) choose $g(\cdot)$ by optimizing the below optimization problem:
	\begin{small}
	\begin{equation}
		\label{boundedsd}
		g^*=\arg\min_{g}\sum_{j=1}^{m}\sum_{i=1}^{n}\mathbb{E}_{s\sim\mathcal{S}^*,u\sim\mathcal{U}^*}\big\lVert SD(s(\mathbf{x}^{(j,i)},k), u(\mathbf{x}^{(j,i)},T))\big\rVert_2^2
	\end{equation}
	\end{small}
\end{proof}
\vspace{-1.0em}

The Lemma~\ref{boundedloss} suggests that the augmented time-series generated by scalable operation $s\sim\mathcal{S}$ and non-scalable operation $u\sim\mathcal{U}$ should exhibit small spectral distance. However, in Alg.~\ref{alg:scalable}, segmenting the time-series into $K$ non-overlapping subseries may disrupt the $\lfloor\frac{T}{K}\rfloor$ lowest frequency components. Therefore, we define the linear function $g(\cdot)$ to extract the lowest $\lfloor\frac{T}{K}\rfloor$ frequency components, thereby preserve such vulnerable patterns. The $k$-th value of function $g(\cdot)$ can be formulated as:
\begin{equation}
	g(\mathbf{x}^{(j,i)})[k] = \sum_{h=0}^{\lfloor\frac{T}{K}\rfloor}2\cdot amp_{h}\cdot\cos(2\pi f_h(k-1)+\phi_h)
\end{equation} 
\noindent where $h=1,2,\cdots,T$, $amp_h$ is the amplitude, $f_h$ is the angular frequency, and $\phi_h$ is the phase of the $h$-th lowest frequency component.

\textit{2) \textbf{Missing values}}. Given a minibatch $\{\mathbf{X}^{(j)}\}_{j=1}^{m}$ of time-series data, which may contain missing values requiring imputation, we are motivated by the recognition that the low-frequency components carry essential information, while the high-frequency components often introduce noise. To address this, we initially employ linear interpolation to handle missing values in $\mathbf{X}^{(j)}$. Subsequently, to filter out the high-frequency noise introduced by linear interpolation, we  apply a moving average with a window size of $10$.

\subsubsection{\textbf{Training algorithm}}
We summarize the learning objectives and scalable algorithm of the data augmentation module in Algorithm~\ref{alg:data_augmentation}. To calculate the expectations in Eq.\ref{eq:preservation_loss} and Eq.\ref{eq:diversity_loss}, we introduce two hyper-parameters, denoted as $C_1$ and $C_2$.   
Here, $C_1$ represents the number of samplings for the unified operations in the spectrum-preservation loss (line 8), and $C_2$ is the number of sampled operation pairs in the spectrum-diversity loss (line 14), where $C_2\leq\binom{C_1}{2}$.  
 The outcomes of parameter-sensitive experiments are presented in Sec.~\ref{sensitivity}. Empirically, employing small values for $C_1$ and $C_2$ has demonstrated both efficiency and effectiveness.

\textbf{Time complexity}. 
The time complexity of Alg.~\ref{alg:data_augmentation} is analyzed as follows. Assuming the length of each time-series is denoted as $T$, the loop from lines 2 to 16 takes $O(m)$ time to traverse all input time-series. Lines 3 to 4 perform linear interpolation and moving average with a small constant window size, which takes $O(T)$ time. Line 8 involves a scalable operation comprising matrix multiplication and addition operations based on Eq.~\ref{matrix_form} and Eq.~\ref{sclable algorithm}. Let the fixed window length be denoted as $K$, then the time complexity is $O(\lfloor\frac{T}{K}\rfloor\cdot n\cdot (4K^2 +3K))\approx O(T\cdot n\cdot K)$ according to Alg.~\ref{alg:scalable}. Line 10 can be computed using the fast Fourier transform (FFT), which takes $O(n\cdot T\cdot\log T)$ time. Line 13 is not counted since $C_1$ is a small constant. Line 14 reuses the results from line 8 and aims to calculate the $JS$ divergence, which takes $O(n\cdot T)$ time.
Thus, the total time complexity of Algorithm \ref{alg:data_augmentation} is $O(m T+ mnC_1\cdot T \cdot K + mnC_1\cdot T\log T + mnC_2\cdot T)$. In practice, as $m$ represents the number of time-series and $n$ represents the number of variables, both can be organized into a single matrix to leverage GPU resources for accelerated computation.

 	\begin{algorithm}[t]
 	\KwIn{A batch $D_{batch}=\{\mathbf{X}^{(j)}\}_{j=1}^m$, the scalable augmentation family $\mathcal{S}$, number of samplings $C_1$ and $C_2$, where $C_2\leq\binom{C_1}{2}$}
 	\KwOut{The batch loss $\mathcal{L}_{aug}$}
 	$\mathcal{L}_{aug}\leftarrow 0$\\
 	\For{$\mathbf{each\ } \mathbf{X}^{(j)}\in D_{batch}$ }{
 					\If{$\text{has missing}$}
 		{
 			$\mathbf{X}^{(j)}\leftarrow Moving\_Average(Interpolation(\mathbf{X}^{(j)}))$}
 		$l_p,l_d,\text{aug\_list}\leftarrow0,\phi$\\
 		\For{$k=1$ \textbf{to} ${C_1}$}{
 			$\text{draw scalable operation }s^{(k)}\sim\mathcal{S}$\\
 				$\hat{\mathbf{X}}^{(j)} \leftarrow{s}^{(k)}(\mathbf{X}^{(j)})$ // Matrix form Eq.~\ref{matrix_form}\\
 				$\text{aug\_list}[k]$ $\leftarrow$ $\hat{\mathbf{X}}^{(j)}$\\
 			$l_p\leftarrow l_p+\frac{1}{n}\sum_{i=1}^{n}\big\lVert
 			|\mathcal{F}\small(\hat{\mathbf{x}}^{(j,i)}\small))|-
 			|\mathcal{F}\small(\mathbf{x}^{(j,i)}\small)|
 			\big\rVert_2^2$
 	}
 	$l_p \leftarrow l_p/C_1$\\
 	\tcp{Sampling for the diversity loss}
 	\For{$k=1$ \textbf{to} ${C_2}$ }{
 	\tcp{Sampling two indices $\binom{|C_1|}{2}$}
 	$(idx_1,idx_2)\leftarrow Sampling\_pairs(C1)$\\
 	$l_d \leftarrow	l_d + \frac{1}{n}\sum_{i=1}^{n}-\log
 	JS\big(
 	\boldsymbol{\alpha}\odot P(\text{aug\_list}[idx_1][i])
 	||
 	\boldsymbol{\alpha}\odot P(\text{aug\_list}[idx_2][i])
 	\big)$
 }
 	 $l_d \leftarrow l_d/C_2$\\
 	 $\mathcal{L}_{aug} = \mathcal{L}_{aug} + l_p + \lambda\cdot l_d$
}
 \textup{\textbf{Return\ }}
$\frac{1}{m}\mathcal{L}_{aug}$\\
 \caption {The training algorithm of the data augmentation module by using scalable algorithm.}
 \label{alg:data_augmentation}
 \end{algorithm}
\vspace{-1.0em}

\subsection{Pre-train LLM}

\subsubsection{Encoder design} The encoder $f_{\theta}$ comprises two primary components: an input embedding layer and a stack of causal transformer-encoder blocks.  Initially, the input embedding layer partitions the input positive views $\tilde{\mathbf{X}}^{(j)}$ and $\hat{\mathbf{X}}^{(j)}$ generated by the augmentation module into a series of consecutive patches, which may be overlapping or non-overlapping~\cite{Yuqietal-2023-PatchTST}, each with a length of $L_p$. The total count of patches is $P=\lfloor\frac{(T-L_p)}{S}\rfloor+1$, where $S$ represents the sliding stride. 
Subsequently, a reprogramming layer~\cite{liu2024taming} is employed to transform the $\tilde{\mathbf{X}}_p^{(j)}, \hat{\mathbf{X}}_p^{(j)}\in\mathbb{R}^{n\times P\times L_p}$ into the high-dimensional space $\tilde{\mathbf{H}}_p^{(j)}, \hat{\mathbf{H}}_p^{(j)}\in\mathbb{R}^{n\times P\times D}$.
Then, both $\tilde{\mathbf{H}}_p^{(j)}$ and $\hat{\mathbf{H}}_p^{(j)}$ are fed into the causal transformer-decoder blocks for representation learning, which mirror the architecture of the text encoder in ViT-G/14 CLIP, comprising 32 transformer blocks with 1280 hidden dimensions. The weights of these 32 blocks are initialized from ViT-G/14 CLIP as well.
ViT-G/14 CLIP is a large pre-trained model designed for processing visual and textual data, trained on approximately $400M$ pairs of text and images using contrastive learning. We opt for ViT-G/14's text encoder structure due to our shared contrastive-based pre-training paradigm and mutual need for sequence modeling. 
The output of the encoder is the patch-wise embeddings, denoted as $\tilde{\mathbf{Z}}^{(j)},\hat{\mathbf{Z}}^{(j)}\in\mathbb{R}^{n\times P\times D}$. In summary, the encoder can be represented as a mapping function $f_{\theta}: \mathbb{R}^{n\times T}\rightarrow \mathbb{R}^{n\times P\times D}$.

\subsubsection{Contrastive Loss} 
We leverage the hierarchical contrastive loss $\mathcal{L}_{cl}$ introduced by TS2Vec~\cite{yue2022ts2vec}, which contrasts learned embeddings at different scales, thereby  enhancing fine-grained representation learning for a variety of downstream tasks.

\subsubsection{Pre-training paradigm} In the large-scale pre-training, the encoder $f_{\theta}$ is trained in a wide range of time-series data from various domains, denoted as $D_{train}=\{D_1,D_2,\cdots,D_N\}$. However, training sequentially (\textit{i.e.}, $D_1\rightarrow D_2\rightarrow\cdots D_N$) may lead to catastrophic forgetting problem~\cite{goodfellow2013empirical}. To overcome this problem, we preprocess all the time-series data into batches, $D_i\rightarrow\{batch_1^{(i)},batch_2^{(i)},\cdots\}$, where each batch from different domains contain varying length of time-series data. During training, we shuffle batches across all domains to improve the robustness and generality of the learned encoder. 
Instead of optimizing the augmentation module and encoder simultaneously, We train alternatively as described in Alg.~\ref{alg:pretrain_llm}. Such approach can effectively tackle the complex optimization problems and avoid problems such as asynchronous optimization. Specifically, for each batch of data, we iteratively train the data augmentation module for $E_1$ iterations, followed by training the encoder for $E_2$ iterations with the augmentation module held fixed.

The time complexity of Alg.~\ref{alg:pretrain_llm} is $O(N\cdot E_1\cdot E_2)$, and the results of parameter-sensitive experiments are presented in Sec.~\ref{sensitivity}. Empirically, employing small values for $E_1$ and $E_2$ has demonstrated both efficiency and effectiveness.

	\begin{algorithm}[t]
	
	\KwIn{Shuffled datasets $D_{train}=\{D_{batch}^{(k)}\}_{k=1}^{N}$, the scalable augmentation family $\mathcal{S}$, the encoder $f_\theta$, epoch number $E_1$, $E_2$, and $E_3$, learning rate $\eta_1$ and $\eta_2$}
	\KwOut{ The encoder $f_{\theta^*}$ and the decoder $q_{\varphi^*}$}
	\For{$i=1$ \textbf{to} $E_1$}{
		\For{$\mathbf{each\ } D_{batch}\in D_{train}$ }{
			\For{$i=1$ \textbf{to} $E_2$}{
			\tcp{Data augmentation module in Alg.~\ref{alg:data_augmentation}}
			$\mathcal{L}_{aug} \leftarrow scalable\_algorithm(D_{batch},\mathcal{S})$\\
			$\theta_{\mathcal{S}}\leftarrow \theta_{\mathcal{S}} - \eta_1 \triangledown_{\theta_{\mathcal{S}}}(\mathcal{L}_{aug})$
			}
		\For{$i=1$ \textbf{to} $E_3$}{
			\tcp{Updates encoder $f_{\theta}$}
			$\mathcal{L}_{cl} \leftarrow f_{\theta}(D_{batch})$\\
			$\theta_{f}\leftarrow \theta_{f} - \eta_2 \triangledown_{\theta_{f}}(\mathcal{L}_{cl})$
		}
	}
}
\textup{\textbf{Return\ }}
the trained operations $\mathcal{S}^*$ and trained encoder $f_{\theta^*}$
\caption{The pre-training paradigm of UniTS. }
\label{alg:pretrain_llm}

\end{algorithm}

\section{Experiments}\label{sec:exp}
\begin{table*}[t]
 
	\caption{Time-series forecasting tasks. The results are averaged from 4 different prediction lengths, that is $\{24, 36, 48, 60\}$ for ILI and $\{96, 192, 336, 720\}$ for the others. We bold the best performance among LLM-based models, which is on the left-hand side of the two vertical lines. We highlight the  best performance for the entire row by both bolding and underlining it.} 
	\label{tab:forecasting}
	\small
	\centering
 
	\begin{threeparttable}
		\setlength\tabcolsep{4.3pt}
		\begin{tabular}{c|c c|c c|c c|c c|c c||c c|c c|c c|c c}
			\toprule 

			\multirow{2}*{Methods} & \multicolumn{2}{c}{UniCL} &\multicolumn{2}{c}{TimeLLM$\dagger$} &\multicolumn{2}{c}{GPT4TS} & \multicolumn{2}{c}{LLaTA$\dagger$} & \multicolumn{2}{c||}{UniTime$\dagger$} &\multicolumn{2}{c}{TimesNet} &\multicolumn{2}{c}{PatchTST} &\multicolumn{2}{c}{LightTS} &\multicolumn{2}{c}{DLinear}  \\
			& MSE & MAE & MSE & MAE & MSE & MAE & MSE & MAE & MSE & MAE & MSE & MAE & MSE & MSE & MAE & MSE & MAE & MSE \\
			\midrule
			ETTm1 & \uline{\textbf{0.343}} & \uline{\textbf{0.372}} & 0.357 & 0.380 & 0.352 & 0.383 & 0.364 & 0.379 & 0.372 & 0.386 & 0.400 & 0.406 & 0.351 & 0.387 & 0.435 & 0.437 & 0.357 & 0.378  \\
			ETTm2 & 0.271 & \textbf{0.325} & 0.269 & 0.335 & \textbf{0.266} & 0.326 & 0.278 & 0.334 & 0.283 & 0.339 & 0.291 & 0.333 & \uline{\textbf{0.255}} & \uline{\textbf{0.315}} & 0.409 & 0.436 & 0.267 & 0.334 \\
			ETTh1 & \textbf{0.422} & \textbf{0.420} & 0.432 & 0.435 & 0.427 & 0.426 & 0.425 & 0.423 & 0.435 & 0.429 & 0.458 & 0.450 & \uline{\textbf{0.413}} & \uline{\textbf{0.430}} & 0.491 & 0.497 & 0.423 & 0.437  \\
			ETTh2 & \textbf{0.334} & \uline{\textbf{0.378}} & 0.349 & 0.393 & 0.346 & 0.394 & 0.342 & 0.389 & 0.352 & 0.395 & 0.414 & 0.427 & \uline{\textbf{0.330}} & 0.379 & 0.602 & 0.543 & 0.431 & 0.447  \\
			ECL & \uline{\textbf{0.160}} & 0.257 & 0.173 & 0.269 & 0.167& 0.263  & 0.162 & \textbf{0.256} & 0.181 & 0.282  & 0.192 & 0.295 & 0.161 & \uline{\textbf{0.253}} & 0.229 & 0.329 & 0.166 & 0.263  \\
			Weather  & \textbf{0.227} & \uline{\textbf{0.264}} & 0.245 & 0.277 & 0.237 & 0.270 & 0.231 & 0.273 & 0.245 & 0.279 & 0.259 & 0.287 & \uline{\textbf{0.225}} & 0.264 & 0.261 & 0.312 & 0.249 & 0.300  \\
			ILI  & \textbf{1.912} & \textbf{0.891} & 1.937 & 0.915 & 1.925 & 0.903 & 1.931 & 0.908 & 2.270 & 1.040 & 2.139 & 0.931 & \uline{\textbf{1.443}} & \uline{\textbf{0.798}} & 7.382 & 2.003  & 2.169 & 1.041  \\
			\midrule
			Average  & \textbf{0.524} & \textbf{0.415} & 0.537 & 0.429 & 0.531 & 0.424 & 0.533 & 0.423 & 0.591 & 0.450 & 0.593 & 0.447 & \uline{\textbf{0.454}} & \uline{\textbf{0.404}} & 1.401 & 0.651 & 0.573 & 0.457  \\
			\bottomrule
		\end{tabular}
		\begin{tablenotes}    
			\footnotesize               
			\item $\dagger$ means that we modify the official code (\textit{e.g.}, modify the input embedding layer) for fair comparison.       
		\end{tablenotes}         
 
	\end{threeparttable}
\end{table*}

We evaluate our approach across a diverse set of time-series analysis tasks, as outlined in Sec.~\ref{sec:preliminary}, encompassing time-series forecasting and time-series classification. Our comparative study  encompasses a comprehensive range of state-of-the-art models, including supervised models and foundational time-series models. Moreover, we conduct additional ablation analyses to assess the scalability and efficacy of the proposed data augmentation methods.

\subsection{Experiment Settings}\label{ssec:exp:setting}
\subsubsection{Datasets} 

In our study, we assess the effectiveness of our model, UniCL, across a diverse array of datasets for various time-series analysis tasks. To ensure fair comparison, we adhere to the experimental settings established in TimesNet~\cite{wu2023timesnet}. 
For time-series forecasting, we utilize seven well-established real-world datasets:  four ETT datasets~\cite{zhou2021informer} (ETTh1, ETTh2, ETTm1, ETTm2), Weather\footnote{\url{https://www.bgc-jena.mpg.de/wetter/}}, Electricity\footnote{\url{https://archive.ics.uci.edu/ml/datasets/ ElectricityLoadDiagrams20112014}}, and ILI\footnote{\url{https://gis.cdc.gov/grasp/fluview/fluportaldashboard.html}}. 
For time-series classification, we evaluate the performance across ten multivariate UEA classification datasets~\cite{bagnall2018uea}.

\subsubsection{Baselines}
We thoroughly compare our approach against a wide-range of time-series models, including the followings: 
(1) Transformer-based supervised models:   ETSformer~\cite{woo2022etsformer}, FEDformer~\cite{zhou2022fedformer}, PatchTST~\cite{Yuqietal-2023-PatchTST}, Autoformer~\cite{wu2021autoformer}, Informer, Non-stationary Transformer~\cite{liu2022non}, and Flowformer~\cite{huang2022flowformer} ; 
(2) Other supervised models: TimesNet~\cite{wu2023timesnet}, LightTS~\cite{zhang2022less}, DLinear~\cite{zeng2023transformers}, TCN~\cite{franceschi2019unsupervised}, XGBoost~\cite{chen2016xgboost}, LSTNet~\cite{lai2018modeling}, and Rocket~\cite{dempster2020rocket} ; 
(3) LLM-based time-series foundation models: Time-LLM~\cite{jin2023time}, GPT4TS~\cite{zhou2023onefitsall}, UniTime~\cite{liu2023unitime}, and LLaTA~\cite{liu2024taming}.

\subsubsection{Implementation details}

With the aid of the scalable algorithm, UniCL undergoes initial pre-training on 40 cross-domain datasets sourced from the Monash Time Series Forecasting Repository~\cite{godahewa2021monash}. These datasets encompass data with varying lengths, ranging from $2$ to $7M$, and include instances with missing values, with missing rate ranging from $2\%$ to $17\%$.  It should be notice that the training data in the experiments is not involved.
We pre-train our model on four NVIDIA A800 GPU for one week. 
Specifically, the ViT-G/14 Clip encoder utilizes a layer count of $32$, with attention blocks fine-tuned using LoRA~\cite{hu2021lora}. Other parameters are also fine-tuned during pretraining. The fixed window size $k$ of the scalable algorithm is set to $120$, while $E_1$ and $E_2$ in Alg.\ref{alg:pretrain_llm} are configured as $3$ and $1$ respectively. The values of $C_1$ and $C_2$ in Alg.~\ref{alg:data_augmentation} are chosen as $5$ and $2$ respectively, and $\lambda$ in Eq.~\ref{eq:summary_loss} is set to $0.01$. Both the augmentation module and the encoder are trained using the AdamW optimizer with an initial learning rate of $0.0001$.
After pre-training, we integrate our trained encoder into the existing pipeline\footnote{\url{https://github.com/thuml/Time-Series-Library}} for evaluation. We also adhere to the same experimental settings as in Wu et al.~\cite{wu2023timesnet} for all tasks.

\subsection{Main Experiments}


\begin{figure}[h] 
	\centering 
	\includegraphics[width=0.5\textwidth]{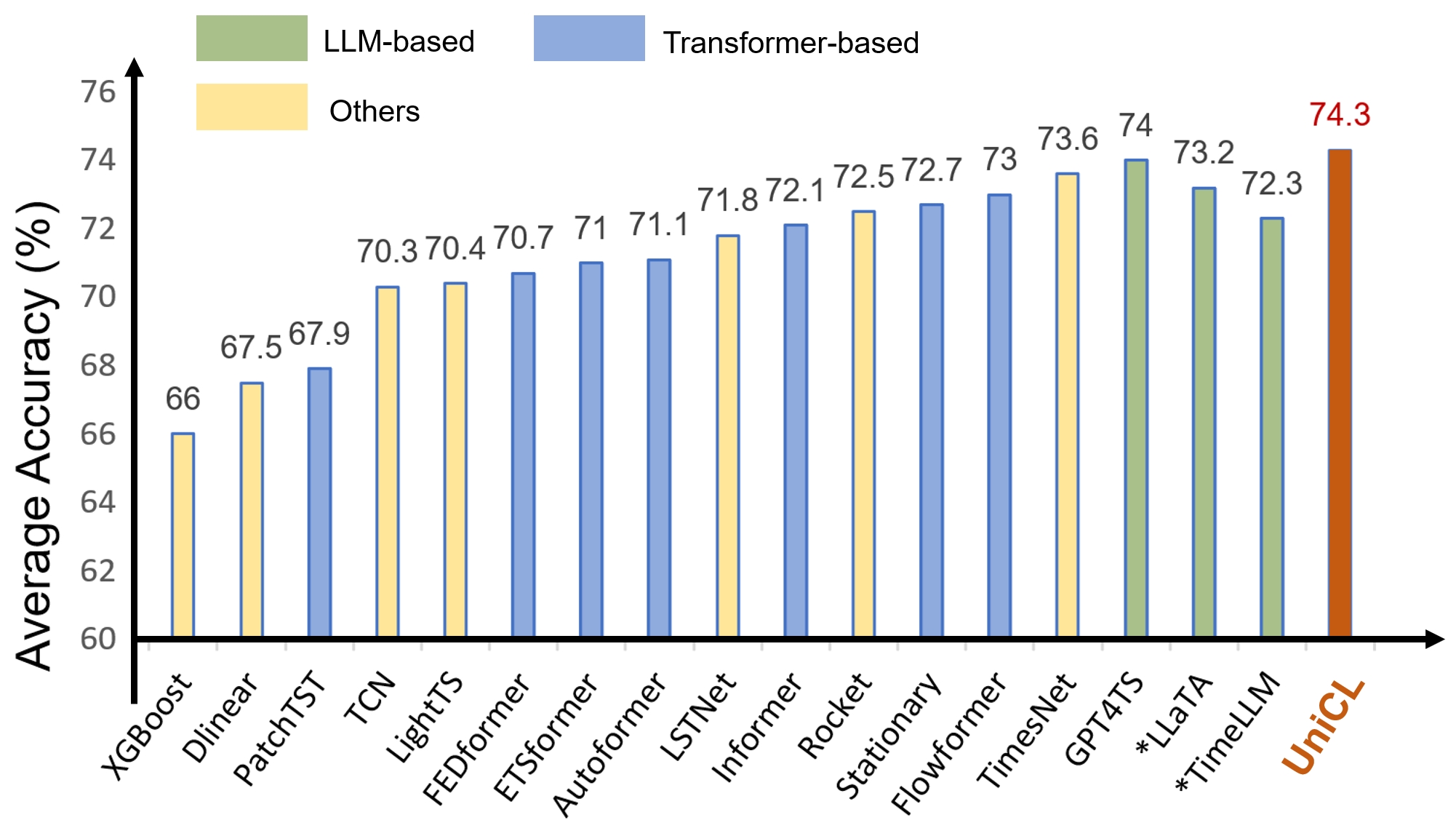} 
	\captionsetup{skip=3pt}
	\captionsetup{justification=justified,singlelinecheck=false,font={small,stretch=0.9}}
	\caption{Results of time-series classification tasks. The accuracy is averaged from 10 subsets of UEA. $*$ means that we modify the baseline's code (\textit{e.g.}, modify the output layer for classification). Other results are from GPT4TS~\cite{zhou2023onefitsall}.} 
	\label{fig:classification} 
 \vspace{-1.5em}
\end{figure}

\subsubsection{Time-series forecasting}
Time-series forecasting tasks involve prediction horizons ranging from $24$ to $720$, with evaluation metrics including mean square error (MSE) and mean absolute error (MAE). 
 Tab.~\ref{tab:forecasting} presents a comprehensive overview of the experimental findings.
 We utilize two vertical lines to demarcate the table. The right part of the table represents state-of-the-art supervised learning models, while the left part pertains to LLM-based time-series foundation models that leverage natural language knowledge for time-series analysis.
UniCL exhibits superior performance across two evaluation criteria, surpassing all LLM-based time-series foundation models. Notably, UniCL outperforms the state-of-the-art LLM-based model GPT4TS, resulting in a notable $1.3\%$ average reduction in MSE and a $2.1\%$ average reduction in MAE. 
UniCL demonstrates competitive results when compared to the supervised models PatchTST, significantly bridging the gap between foundation models and supervised models.


\begin{figure*}[t]
		\vspace{-1.5em}
	\begin{minipage}[t]{0.25\linewidth}
		\centering
		\includegraphics[width=1\textwidth]{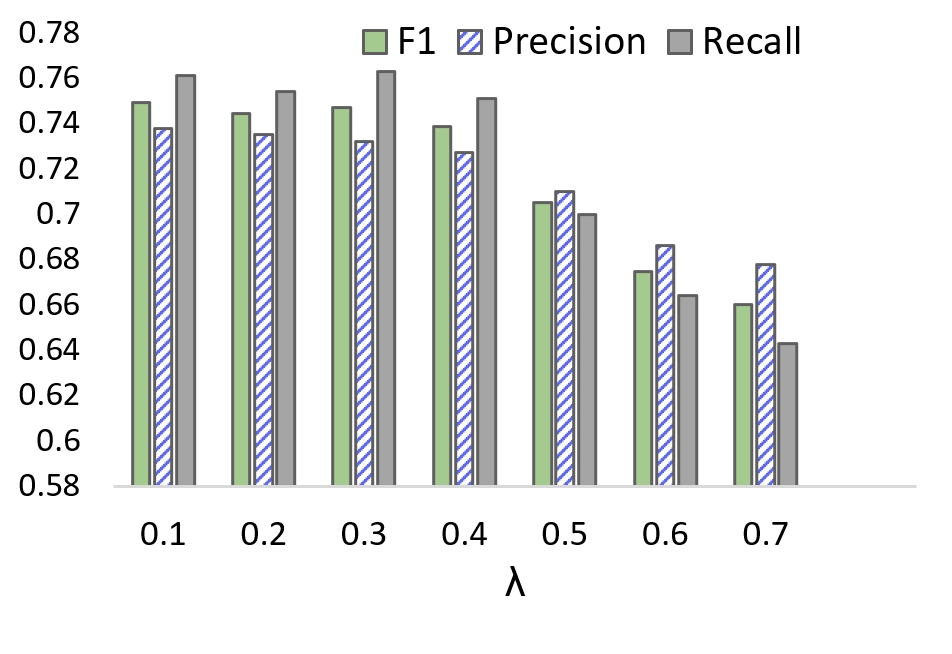}
		\captionsetup{skip=0pt,font={small,stretch=1.2}}
		\caption*{  (a) Effects of $\lambda$}
	\end{minipage}%
	\begin{minipage}[t]{0.25\linewidth}
		\centering
		\includegraphics[width=1\textwidth]{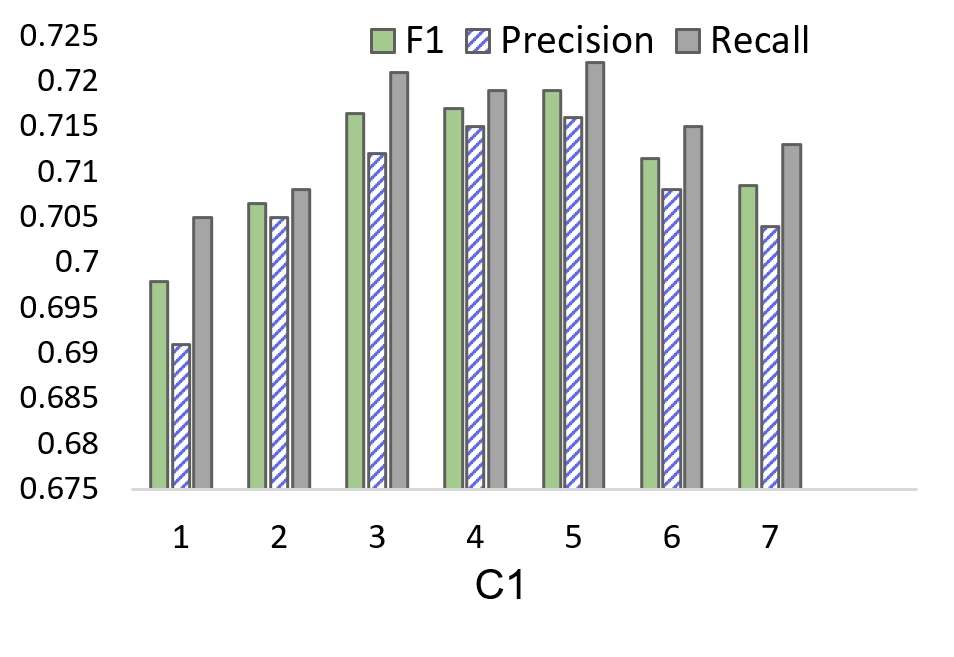}
		\captionsetup{skip=0pt,font={small,stretch=1.2}}
		\caption*{(b) Effects of $C_1$}
	\end{minipage}
	\begin{minipage}[t]{0.25\linewidth}
		\centering
		\includegraphics[width=1\textwidth]{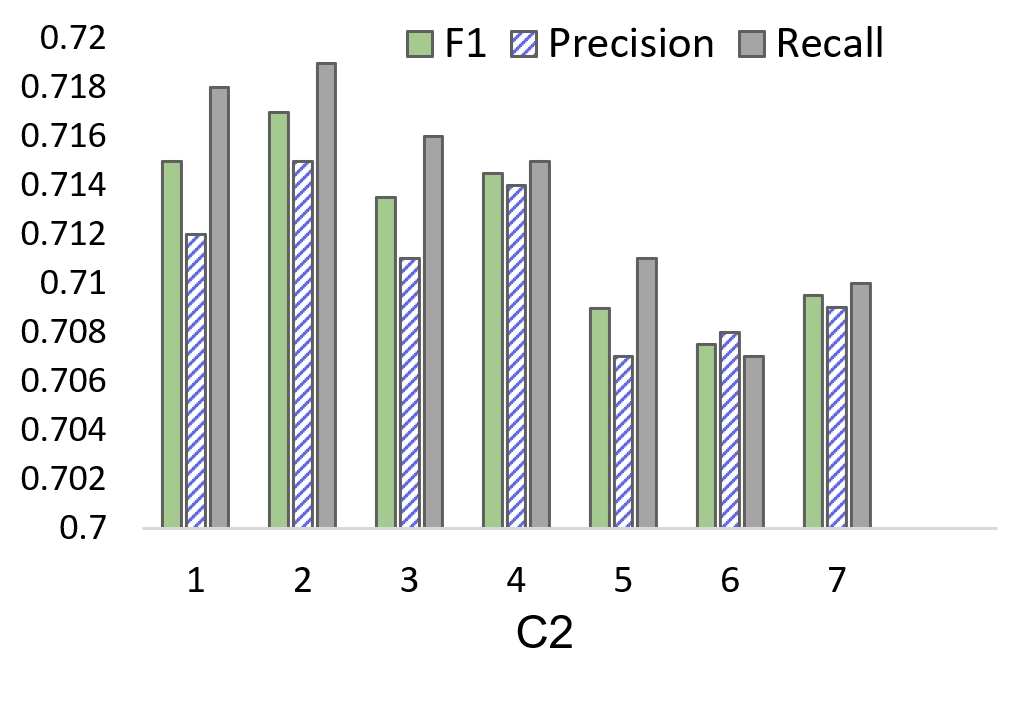}
		\captionsetup{skip=0pt,font={small,stretch=1.2}}
		\caption*{(c) Effects of $C_2$}
	\end{minipage}
	\begin{minipage}[t]{0.24\linewidth}
		\centering
		\includegraphics[width=1\textwidth]{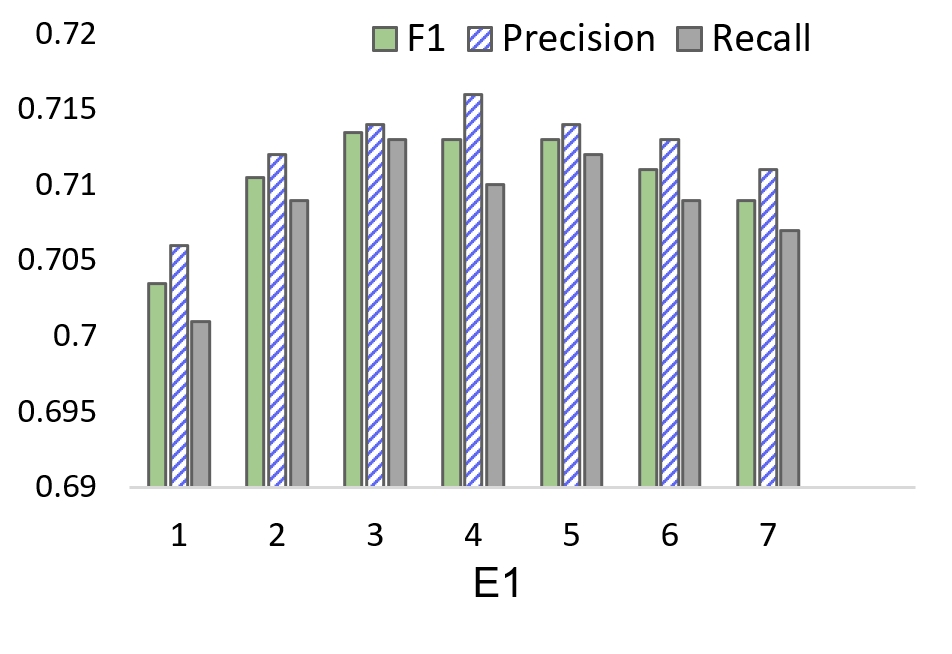}
		\captionsetup{skip=0pt,font={small,stretch=1.2}}
		\caption*{(d) Effects of $E_1$}
	\end{minipage}
	\captionsetup{skip=3pt}
	\captionsetup{justification=justified,singlelinecheck=false,font={small,stretch=0.9}}
	\caption{Parameter sensitivity evaluation. When testing $C_1$, we fix $C_2=2$, and when testing $C_2$, we fix $C_1=3$.}
		\vspace{-1.5em}
	\label{fig:ablation_parameter_sensitive}
\end{figure*}

\subsubsection{Time-series classification} As depicted in Fig.~\ref{fig:classification}, UniCL attains an average accuracy of $74.3\%$, surpassing all baseline methods across 10 multivariate UEA datasets. Notably, UniCL outperforms the state-of-the-art forecasting models TimeLLM and LLaTA by approximately $1\%$, while also surpassing the previous state-of-the-art transformer-based models PatchTST by $5.8\%$.  The superior performance of UniCL may be attributed to the good generality of contrastive learning, which effectively discriminates between diverse and unseen positive and negative views.


%
%
%


\subsection{Ablation Study}

\subsubsection{\textbf{Ablation on the losses and unified operations}}

To better understand the effectiveness of the data augmentation module designs in UniCL, a comparison between full UniCL and its 8 variants on 128 UCR datasets~\cite{dau2019ucr} is shown in Tab.~\ref{tab:ablation_aug}, where 1) \textbf{w/o Spectrum-Preservation Loss} removes the spectrum-preservation loss; 2) \textbf{w/o Spectrum-Diversity Loss} removes the spectrum-diversity loss; 3) \textbf{Unified operation $\rightarrow X$}: replaces our proposed unified and learnable operation into 6 different pre-defined augmentation strategies, including jittering, scaling, permutation, masking, pooling, and warping. 
We observe a significant drop in performance when removing the Spectrum-Preservation Loss, suggesting that by learning spectrum-preserved patterns, the model can effectively capture important patterns within the time-series data.
The absence of Spectrum-Diversity Loss resulted in a notable decrease in accuracy, amounting to $9.2\%$. This underscores the importance of diverse positive views for ensuring the discriminative power of learned embeddings.
Furthermore, decreased performance is observed across all augmentation-based variants, with reductions ranging from $2.1\%$ (masking) to $6.1\%$ (scaling). The relatively minor decrease in performance observed with jittering and masking may be attributed to their ability to generate spectrum-preserved and diverse samples.
In summary, the full UniCL leads to the best performance.

\begin{table}[h]
	\setlength{\abovecaptionskip}{0.2cm}
	\setlength{\belowcaptionskip}{-0.2cm}
	\caption{Ablation study on 128 UCR datasets. The results are averaged from 128 time-series classification tasks.}
	\label{notations}
	\begin{tabular}{llc}
		\toprule 
		&	&  Avg. Accuracy\\ 
		\midrule 
		\hline 
		\multicolumn{2}{c}{\textbf{Full UniCL}} & \textbf{0.844}\\
		\hline
		\multirow{2}*{Loss}  & w/o Spectrum-Preservation Loss  & $0.710(-13.4\%)$\\
		~ & w/o Spectrum-Diversity Loss  & $0.752(-9.2\%)$\\
		\hline
		\multirow{5}*{\makecell[l]{Augment-\\ation}}  & Unified operation $\rightarrow$ Jittering & $0.815(-2.9\%)$\\
		~ & Unified operation $\rightarrow$ Scaling & $0.783(-6.1\%)$\\
		~ & Unified operation $\rightarrow$ Permutation & $0.794(-5.0\%)$\\
		~ & Unified operation $\rightarrow$ Masking & $0.823(-2.1\%)$\\
		~ & Unified operation $\rightarrow$ Pooling & $0.810(-3.4\%)$\\
		~ & Unified operation $\rightarrow$ Warping & $0.818(-2.6\%)$\\
		\specialrule{0cm}{1pt}{2pt}		
		\bottomrule 
	\end{tabular}
	\label{tab:ablation_aug}
\end{table}

\subsubsection{\textbf{Ablation on the scalable algorithm}}

\begin{figure}[h]
	\begin{minipage}[t]{0.5\linewidth}
		\centering
		\includegraphics[width=1\textwidth]{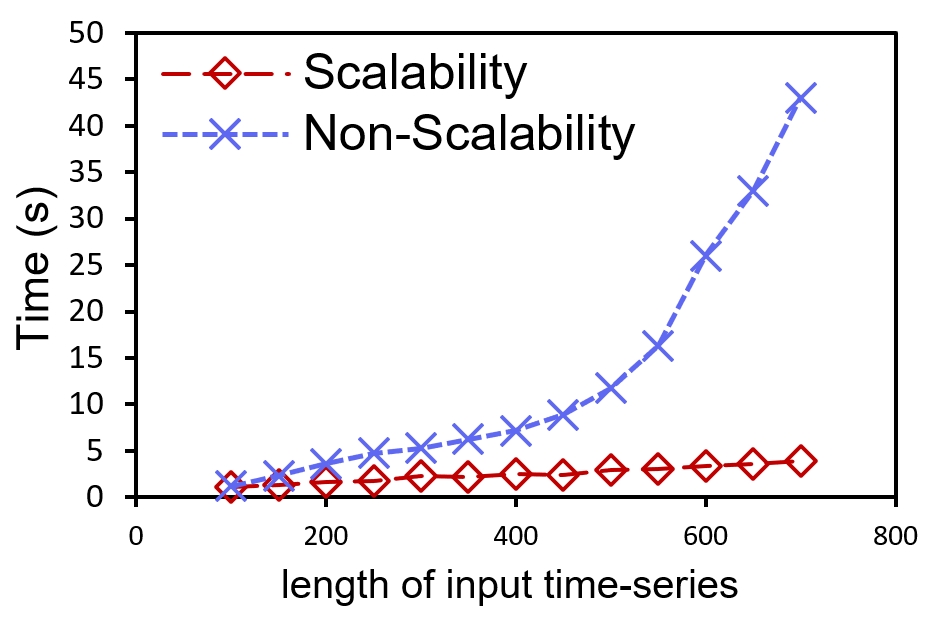}
		\captionsetup{skip=0pt,font={small,stretch=1.2}}
		\caption*{  (a) Computation Time Per Epoch}
	\end{minipage}%
	\begin{minipage}[t]{0.5\linewidth}
		\centering
		\includegraphics[width=1\textwidth]{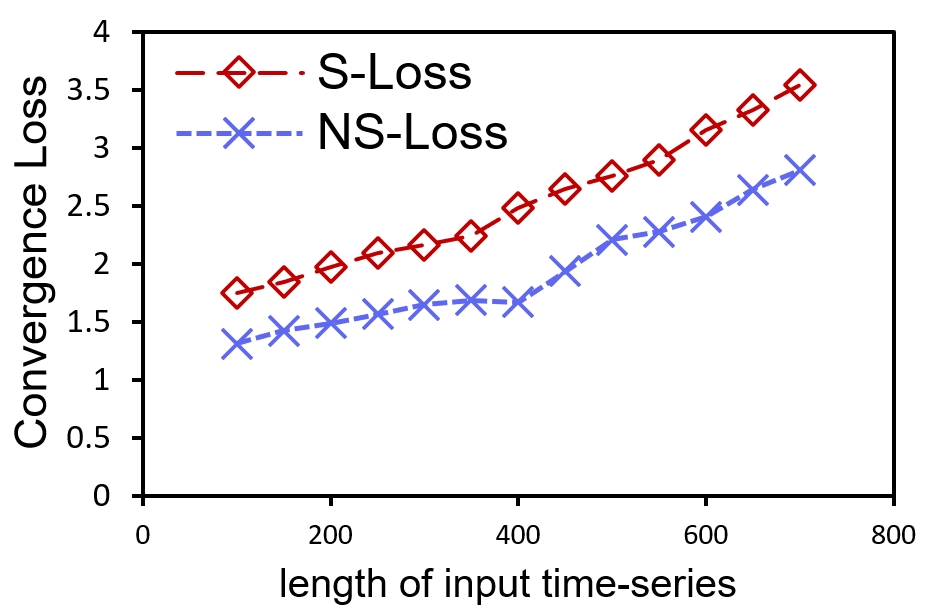}
		\captionsetup{skip=0pt,font={small,stretch=1.2}}
		\caption*{(b) Convergence Loss}
	\end{minipage}
	\captionsetup{skip=3pt}
	\captionsetup{justification=justified,singlelinecheck=false,font={small,stretch=0.9}}
	\caption{Efficiency estimation regarding the length of the input time-series and its effect on the corresponding convergence loss. The window size of the scalable algorithm is set to $100$. S-Loss refers to the convergence loss of the scalable operation and NS-Loss refers to the convergence loss of the non-scalable operation.}
	\label{fig:ablation}
\end{figure}

To better understand the role of the scalable algorithm designs in UniCL, we evaluate the efficiency of the scalable operations and non-scalable operations, as illustrated in Fig.~\ref{fig:ablation}(a). The experiment is conducted on synthetic time-series data with varying lengths, and the time cost per epoch and final average is recorded. 
We observe that beyond a length of $300$, the time cost of non-scalable operations increases exponentially, whereas the time cost of scalable operations remains linear with the length of the input. This is due to the design of the scalable algorithm.
To demonstrate the effectiveness of the scalable algorithm, the corresponding convergence losses of Fig.~\ref{fig:ablation}(a) are illustrated in Fig.~\ref{fig:ablation}(b). Despite the significantly lower time consumption of scalable operations compared to non-scalable ones, we observe that the difference in convergence loss between scalable and non-scalable operations is bounded.

\subsection{Parameter Sensitivity}
\label{sensitivity}

We evaluate the performance of the UniCL based on different parameters: $\lambda$ in Eq.~\ref{eq:summary_loss}, $C_1, C_2$ in Alg.~\ref{alg:data_augmentation}, and $E_1$ in Alg.~\ref{alg:pretrain_llm}. 
The experiment is conducted on two anomaly detection datasets: Yahoo~\cite{yahoo} and KPI~\cite{10.1145/3292500.3330680}, following an evaluation protocol~\cite{10.1145/3292500.3330680}.
As demonstrated in Fig.~\ref{fig:ablation_parameter_sensitive}(a),  the
F1 metric drops with $\lambda$ larger than $0.4$, suggesting that a wise selection of $\lambda$ should be less than $0.4$, which is consistent with Lemma~\ref{boundedloss}. 
According to  Fig.~\ref{fig:ablation_parameter_sensitive}(b,c), the wise selection for $C_1$ and $C_2$ falls within the range of $3$ to $5$ and $1$ to $3$, respectively, underscoring the importance of the spectrum-preserved property over the spectrum-diverse property.
Furthermore, the choice of $E_1$ holds critical importance in alternative training, with the results indicating that $E_1$ should fall within the range of $3$ to $6$. A smaller value of $E_1$ may result in underfitting of the augmentation module.

\vspace{-1.5em}
\section{Conclusion}\label{sec:conclusion}
In this paper, we propose UniCL, a universal contrastive learning framework for pre-training time series models across diverse domains, which addresses the limitations of high bias and low generality in existing models.
First, we propose a novel augmentation operation that preserves spectral properties, ensuring diversity and reduced bias in augmented time-series data.
Second, we propose a scalable augmentation algorithm to time series data with high variability in terms of domain, sequence length, and variable number, thus facilitating robust cross-domain pre-training. 
The superior effectiveness of our
proposed UniCL is demonstrated by extensive experiments
on two benchmarks across eleven domains.

\bibliographystyle{ACM-Reference-Format}
\bibliography{defense_GNN.bib}

\end{document}